\documentclass{article}

\usepackage[final,nonatbib]{neurips_2024}

\usepackage{graphicx}
\usepackage{svg}
\usepackage{amsmath}

\usepackage[utf8]{inputenc} 
\usepackage[T1]{fontenc}    
\usepackage{hyperref}       
\usepackage{url}            
\usepackage{booktabs}       
\usepackage{amsfonts}       
\usepackage{nicefrac}       
\usepackage{microtype}      
\usepackage{xcolor}         
\usepackage{marvosym}

\newcommand{\prac}[2]{\frac{\partial #1}{\partial #2}}

\newcommand{\dih}{d_I(H)}

\title{Scaling Law for Time Series Forecasting}

\author{%
  Jingzhe Shi\textsuperscript{1}\thanks{Equal Contribution.}~, Qinwei Ma\textsuperscript{1}\footnotemark[1]~, Huan Ma\textsuperscript{2},~ Lei Li\textsuperscript{3,4~\Letter}\\
  $^1$ Institute for Interdisciplinary Information Sciences, Tsinghua University\\
  $^2$ Zhili College, Tsinghua University\\
  $^3$ University of Copenhagen $^4$ University of Washington \\
  \texttt{$^{1,2}$ \{shi-jz21,mqw21,mah21\}@mails.tsinghua.edu.cn, $^{3,4}$ lilei@di.ku.dk}\\
}

\begin{document}

\maketitle

\begin{abstract}

  Scaling law that rewards large datasets, complex models and enhanced data granularity has been observed in various fields of deep learning. Yet, studies on time series forecasting have cast doubt on scaling behaviors of deep learning methods for time series forecasting: while more training data improves performance, more capable models do not always outperform less capable models, and longer input horizons may hurt performance for some models.
  We propose a theory for scaling law for time series forecasting that can explain these seemingly abnormal behaviors. 
  We take into account the impact of dataset size and model complexity, as well as time series data granularity, particularly focusing on the look-back horizon, an aspect that has been unexplored in previous theories.
  Furthermore, we empirically evaluate various models using a diverse set of time series forecasting datasets, which (1) verifies the validity of scaling law on dataset size and model complexity within the realm of time series forecasting, and (2) validates our theoretical framework, particularly regarding the influence of look back horizon. We hope our findings may inspire new models targeting time series forecasting datasets of limited size, as well as large foundational datasets and models for time series forecasting in future work.\footnote{Code for our experiments has been made public at: \url{https://github.com/JingzheShi/ScalingLawForTimeSeriesForecasting}.}

\end{abstract}

\section{Introduction}

Because of the practical value of time series forecasting, past years have seen rapid development for time series forecasting methods using the paradigm of neural network training. Neural Nets utilize different model architectures, including FFN-based~\cite{dlinear,tsmixer,xu2024fits}, Transformer-based~\cite{informer,autoformer,fedformer,Yuqietal-2023-PatchTST,itransformer} and Convolution-based~\cite{timesnet,moderntcn} neural nets have been proposed. Starting from around 2022, some previous work~\cite{dlinear,micn,Yuqietal-2023-PatchTST,moderntcn} proposed that powerful models could be enhanced by extending the look-back horizon because more historical information can be utilized.
However, our experiments (Figure \ref{fig:HorizonXData_diffDatasets}) show that \textbf{this claim may not hold for datasets in practice} \textbf{with a certain amount of training data}: optimal horizon does exist, and it will increase if the amount of available training data increases. This calls for a more thorough understanding of the impact of horizon and dataset size on forecasting loss.

In Natural Language Processing (NLP)~\cite{openaiscaling,chinchillascaling}, Computer Vision (CV)~\cite{visionscaling} and other fields in deep learning, the impact of dataset size, model size and data granularity on performance is sometimes summarized as the Scaling Law: larger dataset, larger models and more detailed data granularity improves performance in these cases, and theories~\cite{explainingneural,manifoldscaling} have been proposed to explain these behaviors. \textbf{However, these theories do not lay emphasis on the horizon of time series}, hence cannot be used directly to explain the impact of the horizon.

In this work, we introduce a comprehensive scaling law theory for time series forecasting, with a particular emphasis on the impact of the horizon. This theory integrates dataset size and model size to optimize predictive performance based on the look-back horizon. We further conduct experiments to (1) verify that scaling behaviors for dataset size and model size do exist in time series forecasting and (2) validate our theory, especially about the influence of different horizons.

Our main contribution regarding time series forecasting includes:
\begin{itemize}
\item[1.] We introduce a novel theoretical framework that elucidates scaling behaviors from an intrinsic space perspective, highlighting the critical influence of the look-back horizon on model performance. Our theory identifies an optimal horizon, demonstrating that beyond this point, performance degrades due to the inherent limitations of dataset size.
\item[2.] We conduct a comprehensive empirical investigation into the scaling behaviors of dataset size, model size, and look-back horizon across various models and datasets. Our research establishes a robust scaling law for time series forecasting, providing a foundational framework that adapts to diverse modeling contexts.
\end{itemize}

As a corollary to our conclusions, we point out that different models might have different optimal look-back horizon for the same dataset (Figure \ref{fig:DataScaling}); therefore, \textbf{we call for future work to compare different models using the optimal look-back horizon for each model correspondingly rather than using a fixed look-back horizon}.

As a further result of our findings, though widely used in previous work, \textbf{to show a model benefits from longer horizon compared to baseline models is unnecessary for proving its superiority over these baseline models}.

We hope our work may inspire future work when designing forecasting models for specific datasets of limited size, as well as future work proposing large foundational datasets and models in the field of time series.


\section{Related Work}\label{Related Work}
\subsection{Time Series Forecasting}

The task of time series forecasting is to predict a time series with $N$ variables in its next $S$ frames (denoted as $Y = \{y_1,y_2,\ldots,y_N\} \in \mathbb{R}^{N\times S}$) given its previous observations with $H$ frames (denoted as $X = \{x_1,x_2,\ldots,x_N\}\in \mathbb{R}^{N\times H}$). $H$ is called look back horizon in some scenarios.

Channel-Independent model means to predict $y_i$ by $\hat{y_i}=f(x_i)$. Linear models and MLPs have been proven to be effective learners for time series forecasting. A series of work~\cite{dlinear,xu2024fits,analysislineartimeseries} utilizes linear layers and methods like low-pass-filter and 
Ordinary Least Squares regression to learn linear relationships in the time series. Reversible MLP~\cite{rmlp} proposes to use linear layers and MLPs with reversible Instance Norm~\cite{revin} and obtains satisfying results. PatchTST~\cite{Yuqietal-2023-PatchTST} proposes to use patch embedding for time series. 

Channel-Dependent model means to predict $y$ by $\hat{y_i}=[f(x_1,x_2,\ldots,x_N)]_i$. A series of works based on Transformers and its variants have been proposed\cite{informer,autoformer,fedformer}, as well as a series of convolution-based methods based on TCN\cite{tcn, tcn_2}. More recently, iTransformer~\cite{itransformer} proposes to use attention to capture the relationship between different variables. ModernTCN ~\cite{moderntcn} proposes new convolutional architectures enabling a wide Effective Receptive Field.

There have been many works analyzing on mathematical properties of time series before machine learning exists. \cite{oldtimetheories,25years5} give good summaries of the early works from different perspectives.

Recently, there have been works proposing large foundational datasets and models for time series. Some propose foundational models that can do zero-shot forecasting\cite{timegpt}. Some propose open-source foundational datasets and verify the ability for transfer learning of foundation models trained on the datasets\cite{liu2024timer,das2024decoderonlyfoundationmodel}.\footnote{Released in close timing with ours, \cite{scalinglawslargetsm} verifies scaling law experimentally for the classic transformer architecture on a large mixed dataset for time series forecasting.} There are also works utilizing LLMs to do zero-shot prediction\cite{llmzeroshottimeseries}, or use LLM backbones pretrained on text contexts to perform time series forecasting\cite{llm4ts,onefitsall,timellm}.







\subsection{Scaling Law and related Theory}
A plethora of research has been conducted to investigate the scaling law in various domains of deep learning, encompassing Natural Language Processing, Computer Vision tasks, and Graph-based Neural Networks~\cite{openaiscaling,chinchillascaling,visionscaling,gnnscaling}. These studies have not only observed the existence of scaling law but also proposed theories to elucidate them. For instance, certain work has interpreted the scaling behavior from the standpoint of intrinsic space or data manifold, providing a deeper understanding of the underlying mechanisms~\cite{explainingneural,manifoldscaling}. In parallel, the properties of time series have been the subject of extensive theoretical and empirical studies. Notably, some research has established bounds for the quantization error of time series, contributing to the body of knowledge on time series analysis~\cite{quantizationerror}. 




\section{Theory for Scaling Law in Time Series Forecasting}\label{Theory for Scaling Laws in Time Seris Forecasting}

\subsection{Forecasting in Intrinsic Space}

To represent the amount of information carried by a time series slice of length $L$, we consult the concept of intrinsic dimension and intrinsic space. Consider all time series slices of length $L$ in a particular scenario, these slices are hypothesized to reside within an intrinsic space that encapsulates the fundamental characteristics of the data, effectively representing its inherent features. Denote this intrinsic space as $\mathcal{M}(L)$ and its intrinsic dimension as $d_I(L)$. 

It immediately follows that \textbf{time series forecasting is equivalent to predicting a vector in} $\mathcal{M}(S)$ \textbf{given its previous $H$ frames, which can be represented by a vector in the space} $\mathcal{M}(H)$. 

\subsection{Formulation} \label{formulation}

Before studying the impact of horizon, dataset size and model on loss, we first formulate the intrinsic space, the data distribution and the form of the loss.

\subsubsection{Intrinsic Space} \label{intrinsic space}

For a time series $s_{0,1,\ldots, L}$ of length $L$ (each $s_i$ is a unit element in the sequence that may contain single or multiple variables, dependent or independent), we represent it using a vector $x$ in $\mathcal M (L)$.  

We make these assumptions on the spaces $\{\mathcal{M}(1),\mathcal{M}(2),\ldots\}$ and the distribution of data in the spaces:
\begin{itemize}
\item[1.] \textbf{Information-preserving}: Intuitively speaking, we should be able to recover the real sequence (which might be a multivariable sequence or a singlevariable sequence) from its corresponding intrinsic vector with the error bounded by a small constant value. Formally we can state this as follows: 

\emph{Exists a mapping $\phi$ from the original length-$L$ sequence space $\mathcal{O}(L)$ to $\mathcal{M}(L)$, an inverse mapping $\phi^{-1}:\mathcal{M}(L)\to \mathcal{O}(L)$ and a constant $e\ll 1$ related to $L$ so that for any $\mathbb{E}_{x \sim \mathcal{O}(L)}$, $\|x-\phi^{-1}(\phi(x))\|_2^2\leq e(L)$}. 
\item[2.] \textbf{Inverse Lipschitz}: $\phi^{-1}$ should be $K_I$-Lipschitz under L2 norm. That is:

\begin{equation*}
    \forall x,y\in\mathcal{M}(L),\|\phi^{-1}(x)-\phi^{-1}(y)\|_2 \leq K_I\|x-y\|_2
\end{equation*}

\item[3.] \textbf{Bounded}: For simplicity, we assume the values in all dimensions of the intrinsic space are bounded, and thus we can scale the intrinsic space to fit it into $\mathcal{M}(H)=[0,1]^{d_I(H)}$.

\item[4.] \textbf{Isomorphism}: $\mathcal M(L_1)$ is isomorphic to a subspace of $\mathcal M(L_2)$ for $L_1\leq L_2$. Moreover, the isometry should also preserve the data distribution in the space. Formally, we can state it as:

\emph{Let $P[H_1,H_2]$ denote the linear projection from $\mathcal{M}(H_1)$ to the subspace of it isomorphic to $\mathcal{M}(H_2)$, and let $Cov_{L}$ denote the covariance matrix of data distribution in $\mathcal{M}(L)$, then $Cov_{L_1}$ should be congruent to $P[L_2,L_1](Cov_{L_2})$ for any $L_1 < L_2$.}

\item[5.] \textbf{Linear Truncation}: Truncation in time series space is close to linear projection mapping in $\mathcal{M}(L)$. 
Formally, we can state it as:

\emph{Define a truncating function $t_p[H_1,H_2](x_{0:H_1})$, so that for any sequence $s$, let $s_{h_1:h_2}$ denote the intrinsic vector for the subsequence from $s[h_1]$ to $s[h_2-1]$, then $t_p[H_1,H_2](s_{0:H_1})=s_{0:H_2}$, then $t_p[H_1,H_2]\approx P[H_1,H_2]$.}

\item[6.] \textbf{Causality}: There should be an optimal model to predict the next $S$ frames given the previous $h\to\infty$ frames, so that the error only originates from the intrinsic noise. That is:

\begin{equation*}
    \exists F[S]: \mathcal{M}\to\mathcal{M}(S), s.t. \lim\limits_{h\to\infty}\mathbb{P}(y\mid x_{-h:0}) = (1-\eta)\delta(F[S](x_{-h:0})) + \eta\mathcal{N}(F[S](x_{-h:0}), \Sigma_S),
\end{equation*}

where $\eta$ stands for the noise ratio in the system, $\delta(\cdot)$ stands for the Dirichlet function and $\mathcal{N}(\mu, \Sigma)$ stands for a normal distribution with mean $\mu$ and covariance $\Sigma$. (Notice that the noise distribution is not necessarily a normal distribution and our result holds for any noise distribution with mean $\mu$ and covariance $\sigma$. However we only consider the normal distribution case here for simplicity.)

Moreover, we assume $F[S]$ is first-order $K_1$-Lipschitz and second-order $K_2$-Lipschitz in $L_2$ metric (as we assume $S$ is fixed we don't discuss how the Lipschitz coefficients vary with $S$ and simply take them as constant).

\item[7.] \textbf{Uniform Sampling Noise}: When drawing a length-$L$ sample, we assume the sampling noise is uniform in each direction in $\mathcal{M}(L)$.

\item[8.] \textbf{Zip-f Distribution}: We assume the data distribution in the intrinsic space follows a Zip-f law on different dimensions of the intrinsic space. That is, the eigenvalue spectrum of $Cov_L$ satisfies $\lambda_i \approx \lambda_0 i^{-\alpha_Z}$ where $\lambda_i$ represents the $i$-th largest eigenvalue. This is shown by other work \cite{levi2024underlying, Petrini_2023} and also verified in our experiments. 

\end{itemize}

This result is asymptotic and does not suggest uniform intrinsic dimensions across sequence elements. Specifically, for a sequence element with $v$ variables, $d_I(H) \propto v$ approximately in channel-independent scenarios. In contrast, for channel-dependent cases, which are more common, the total intrinsic dimension is typically less than the sum of dimensions for individual variables.

In the deduction part we do not assume the specific relationship between $d_I(H)$ and $H$. Some previous work\cite{buzug1995characterising} shows that in some cases $d_I(H) \approx \Theta(H)$. In Appendix \ref{relation d and h} we discuss more about their relationship.

We formulate the intrinsic space and these assumptions more strictly and provide a brief construction of $\{\mathcal{M}(1),\mathcal{M}(2),\ldots\}$ in Appendix \ref{intrinsic construction}. Also, we discuss cases where these conditions are not strictly satisfied in Appendix \ref{non linear trunc}.

Moreover, under these assumptions made, the loss in the original space $L_{ori}$ can be linearly bounded by the loss in the intrinsic space $L_{ins}$, details of which can be found at Appendix \ref{app: reduction of loss into intrinsic space}.

In the following deduction, we use $L$ to denote $L_{ins}$.

\subsubsection{Loss: Overall} \label{loss: overall}
In the following sections, we consider the task to predict a vector in $\mathcal M(S)$ given the vector corresponding to its previous $H$ frames in $\mathcal M(H)$ with a model $m$. For simplicity, we represent the operation of obtaining a vector in \(\mathcal{M}(t_2-t_1)\) by truncating the sequence \(s\) from time \(t_1\) to \(t_2\) as \(x[t_1:t_2]\), where \(x\) is a representation in \(\mathcal{M}(|s|)\).

$$
\begin{aligned}
L&=E_{x\sim\mathcal M(H+S)}[(x[H:H+S]-m(x[0:H]))^2]\\
\end{aligned}
$$
Let $m^*$ denote the optimal Bayesian model, then it should satisfy:
$$
m^*(x[0:H])=E_{x\sim\mathcal M(H+S)}[x[H:H+S]|x[0:H]]
$$
Thus:
$$
\begin{aligned}
    L=&E_{x\sim\mathcal M(H+S)}[(x[H:H+S]-m^*(x[0:H])+m^*(x[0:H]-model(x[0:H]))^2]\\
    =&E_{x\sim\mathcal M(H+S)}[(x[H:H+S]-m^*(x[0:H])^2]+E_{x\sim\mathcal M(H)}[(m^*(x)-m(x))^2]\\
    &+2*E_{x\sim\mathcal M(H+S)}[(x[H:H+S]-m^*(x[0:H])*(m^*(x)-m(x))]
\end{aligned}
$$
Since $x[H:H+S]-m^*(x[0:H])$ and $m^*(x)-m(x)$ are independent with respect to $s[H:H+S]$ and $E_{x\sim\mathcal M(H+S)}[x[H:H+S]-m^*(x[0:H])]=0$, the loss is a sum of the previous two terms: one is decided by the capability of the optimal Bayesian model (or the Bayesian Error), and the other is the model's ability to approximate the Bayes Estimation: $L=L_{Bayesian}+L_{approx}$.

We then calculate each of the two terms corresponding to the horizon, dataset size and model size.

\subsubsection{Bayesian Loss} \label{bayesian loss}

We consider the Bayesian error for $\mathcal M(\infty)$, when the loss for predicting $S$ frames originates from the inherent uncertainty of the system. That is, we evaluate the amount of information carried by $\mathcal M(H)$ compared to $\mathcal M(\infty)$. By assumption 5 and 6 in section~\ref{intrinsic space}, It can be verified that:
$$
    L_{Bayesian} \leq (1-\eta)K_1^2\mathbb{E}[var(P^{-1}[\infty,H](x))] + \eta \cdot tr(\Sigma_S)
$$

According to assumption 7, the noise in the $i$-th predicted frame would be proportional to $\sqrt{i}$, and the total noise in the predicted $S$ frames would be proportional to $S$. Let $\sigma_M^2$ denote the variance of the noise for a single frame in a single dimension, then $tr(\Sigma_S)$ should be equal to $\sigma_M^2S^2d_I(S)$. 

According to assumptions 4 and 8, we can express the inverse projection term into:
$$
\begin{aligned}
\mathbb{E}[var(P^{-1}[H](x))] &= \sum\limits_{d_I(H)\leq i < d_I}\lambda_i \approx \frac{\lambda_0}{(\alpha_Z - 1)d_I(H)^{\alpha_Z-1}}
\end{aligned}
$$
which indicates that:
$$
L_{Bayesian} \approx K_1^2(1-\eta)\frac{\lambda_0}{(\alpha_Z - 1)d_I(H)^{\alpha_Z-1}}
+ \eta\sigma_M^2S^2d_I(S)
$$

\subsubsection{Approximation Loss: Two Cases} \label{approx loss}

The training data is sampled from the distribution of $E_{x\sim\mathcal M(H+S)}[x[H:H+S]|x[0:H]]$. Following previous works \cite{manifoldscaling, pla}, we utilize the piece-wise linear assumption for deep learning models. That is, we assume the model partitions the intrinsic space into $N$ subregions, and does a linear prediction for each subregion independently. Here we discuss two cases: the large-dataset limit and the small-dataset limit.

\textbf{If the dataset is large} and contains many samples in the piece of the model, the model tends to learn the averaged behavior of the region. Intuitively, a larger dataset size brings smaller noise averages, and a larger model brings smaller piece-wise linear pieces, reducing the error caused by deviation of the data distribution to the linear model in the small blocks. Roughly speaking, the loss consists of two terms: one caused by the uncertainly within the subregions partitioned by the model $L_{r}$, and one caused by the noise in the data that makes the model fail to learn the optimal model $L_n$. If we assume that the model partitions the space uniformly, then the loss should satisfy:
$$
L_{approx}\approx K_2^2\frac{\dih^2N^{-\frac 4 {\dih}}}{4\pi^2} + \frac{Nd_I(H)}{D}(\sigma_M^2S^2d_I(S) + \frac{K_1^2\lambda_0}{(\alpha_Z-1)\dih^{\alpha_Z-1}}).
$$
The uniform partitioning should be the most naive case for the model to learn, and we also analyze on the cases where more advanced partitionings are learned in Appendix~\ref{app: partition}. Also, it is worth noticing that the noise consists of two term, one is the systematic uncertainly as shown in section \ref{bayesian loss}, and the other is caused by the horizon limitation so that the effect of unseen dimension seems exactly like noise for the model. Please refer to Appendix~\ref{loss sufficient data} for a more detailed derivation. 

\textbf{Otherwise if there are few data samples, or the model is sufficiently large} and the model cannot learn to average the noises of closed samples, but rather learns to remember each sample in certain pieces. This would give a data-scaling loss determined by nearest-neighbor distance. In this case:
$$
L_{approx}\approx \frac{K_1^2}{4\pi}d_I(H)D^{-\frac 2 {d_I(H)}}.
$$
Please refer to Appendix \ref{loss scarce data} for a more detailed derivation of this scenario.

\textbf{Boundary of the two phases} is decided by Dataset size and Model size, as well as horizon. It is worth noticing that in time series forecasting tasks, the sliding window method for constructing data samples is usually used. \footnote{Data points between time [t-H:t+S] is considered one sample, and [t-H+1:t+S+1] is considered another sample in the dataset.  These two samples are strongly correlated with each other.} Therefore, the closest data points are always dependent on each other and are strongly correlated. The effective number of mutually independent data fragments is approximately proportional to $D/H$ rather than $D$, and $\xi=D/NH$ would be a very approximate order parameter separating the previous two scenarios. Again, please refer to Appendix~\ref{loss deduction} for a more detailed derivation.

\subsection{Optimal Horizon under Each Circumstance} \label{optimal horizon main}
As stated in section~\ref{loss: overall}, the total loss would be simply the sum of the two components deduced above. Analyzing the total loss form, we may achieve an optimal horizon $H^*$ (or a corresponding optimal intrinsic dimension $d_I^*$ that minimizes the loss for different cases.
\subsubsection{Optimal Horizon for large amount of data}
For the case with sufficient data, the total loss is:
$$
\begin{aligned}
    L=&L_{Bayesian}+L_{Approx}\\
    \approx&K_2^2\frac{\dih^2N^{-\frac 4 {\dih}}}{4\pi^2} +  K_1^2(1-\eta)\frac{\lambda_0}{\alpha_Z - 1} \frac{1}{d_I(H)^{\alpha_Z-1}} \\
&+  \frac{Nd_I(H)}{D}(\sigma_M^2S^2d_I(S) + \frac{K_1^2\lambda_0}{(\alpha_Z-1)d_I^{\alpha_Z-1}(H)})
\\
\approx& K_2^2\frac{\dih^2N^{-\frac 4 {\dih}}}{4\pi^2} +  K_1^2(1-\eta)\frac{\lambda_0}{\alpha_Z - 1} \frac{1}{d_I(H)^{\alpha_Z-1}} +  \frac{Nd_I(H)}{D}\sigma_M^2S^2d_I(S)\\
&\text{ (since we always assume $N\ll D$)}
\end{aligned}
$$
We consider two cases. The detailed derivation can be found in Appendix \ref{sufficientdataoptimal}.

\textbf{If model size is too small compared to dataset size} such that $N=o(D^{\frac{d_I(H)}{d_I(H)+4}})$, then the effect of dataset size on picking optimal horizon could be neglected, thus:
$$
d_I^* = \mathcal{W}(\frac{4}{\alpha_Z C_0^{\frac 1 {\alpha_Z}}}\ln^{1+\frac 1 {\alpha_Z}}N)\approx \frac{4}{\alpha_Z C_0^{\frac 1 {\alpha_Z}}}\ln^{1+\frac 1 {\alpha_Z}}N.
$$
where $\mathcal{W}(\cdot)$ is the Lambert W function ($\mathcal{W}(x)\approx x$) and $C_0=\frac{K_1^2\pi^2(1-\eta)\lambda_0}{K_2^2}$.

\textbf{If N is not that small}: $N = \omega(D^{\frac {d_I(H)}{d_I(H)+4}})$. Then the noise effect would be dominant in picking optimal $H$, and the optimal $d_I(H)$ would be:
$$
d_I^* = (\frac{K_1^2(1-\eta)\lambda_0 D}{N\sigma_M^2S^2d_I(S)})^{\frac 1 {\alpha_Z}}.
$$
In this case, $d_I^*$  grows noticeably with the increment of $D$ and the decrement of $N$.
\subsubsection{Optimal Horizon for a relatively small amount of data}
If data is scarce, the loss could be written as:
$$
\text{loss} \approx K_1^2(1-\eta)\frac{\lambda_0}{\alpha_Z - 1} \frac{1}{d_I(H)^{\alpha_Z-1}} + \frac{K_1^2}{4\pi}d_I(H)D^{-\frac 2 {d_I(H)}}
$$
It is worth noticing that in this case, we assume the model is always large enough to find the nearest neighbor for a test sample, hence the loss is irrelevant with $N$. 
We can estimate the optimal $d_I^*$ as:
$$
d_I^* = C_s \frac{\ln D}{\ln\ln D}
$$
Where $C_s$ is a constant irrelevant with $D$, $N$ or $H$. The exact form is provided in Appendix \ref{scarcedataoptimal}. Compared to the first scenario, the optimal $d_I$ changes much less in this scenario.

\section{Experiment Results}\label{Experiment Results}

\subsection{Scaling Law for dataset size and model width do exist for Time Series Forecasting}


\begin{figure*}[h]
  \centering
  \includegraphics[width=\linewidth]{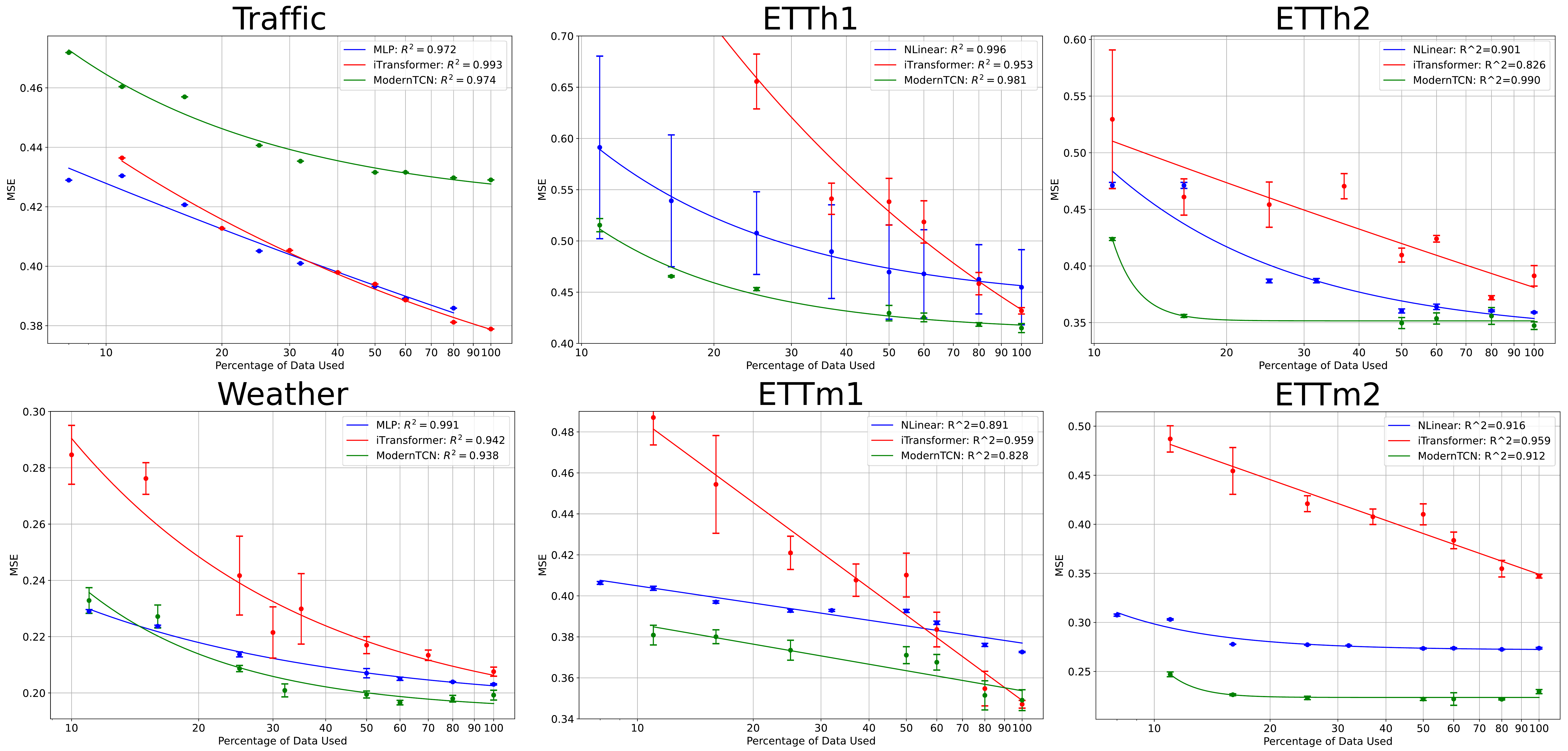}
  \caption{Data Scaling. The proposed formula $loss(D)=A+B/D^\alpha$ fits well. More comparison with other formulas can be found at Appendix \ref{app:other formulas}.}
  \label{fig:DataScaling}
\end{figure*}
\begin{figure*}[h]
    \centering
    \includegraphics[width=1\linewidth]{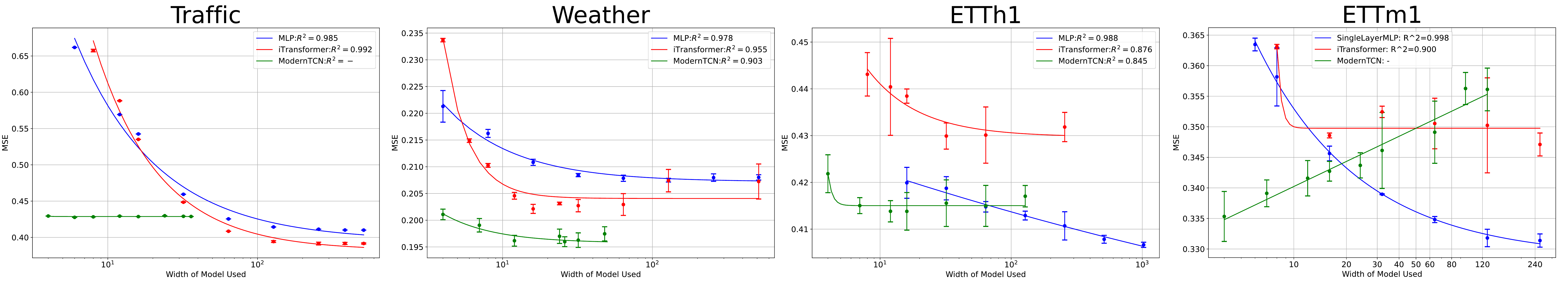}
    \caption{Width Scaling. When the model is not powerful enough, $loss(W)=A+B/W^\alpha$ fits well for these situations. When data is scarce, a large model may lead to overfitting, as observed with ModernTCN on ETTm1.}
    \label{fig:widthScaling}
\end{figure*}
As depicted in Figure~\ref{fig:DataScaling} and Figure~\ref{fig:widthScaling}, we corroborate the scaling behaviors pertaining to data scaling and model-size scaling across a diverse range of datasets and models. This validation underscores the robustness and versatility of our proposed theoretical framework in the context of time series forecasting.

Here we mainly include NLinear\cite{dlinear}/MLP, ModernTCN\cite{moderntcn} and iTransformer\cite{itransformer} as our models, covering a scenario of Channel-Independent and Channel-Dependent, FFN-based, Transformer-based and Convolution-based methods. For datasets, we mainly use ETTh1, ETTh2, ETTm1, ETTm2, Traffic and Weather\cite{informer,autoformer} as our datasets. Detailed experiment settings can be found at Appendix \ref{app:exp}.

It can be seen that for all these models on all these datasets, for the \textbf{dataset-scaling} case where $loss = C_D+1/D^{\alpha_D}$ and the \textbf{model width-scaling} with large amount of data: $loss=C_W+1/W^{\alpha_W}$. The results fit well, thus verifying the existence of the original understanding of the scaling law.

However, in some special cases when the model is large enough to approximate the data, increasing model size would not gain performance, and may hurt performance (if regularization that is not strong enough is added) in some cases, like ModernTCN for ETTm1.

\subsection{The Impact of Horizon to Final Loss}

\subsubsection{Optimal Horizon and Training Data Amount}
\begin{figure*}[!h]
    \centering
    \includegraphics[width=\linewidth]{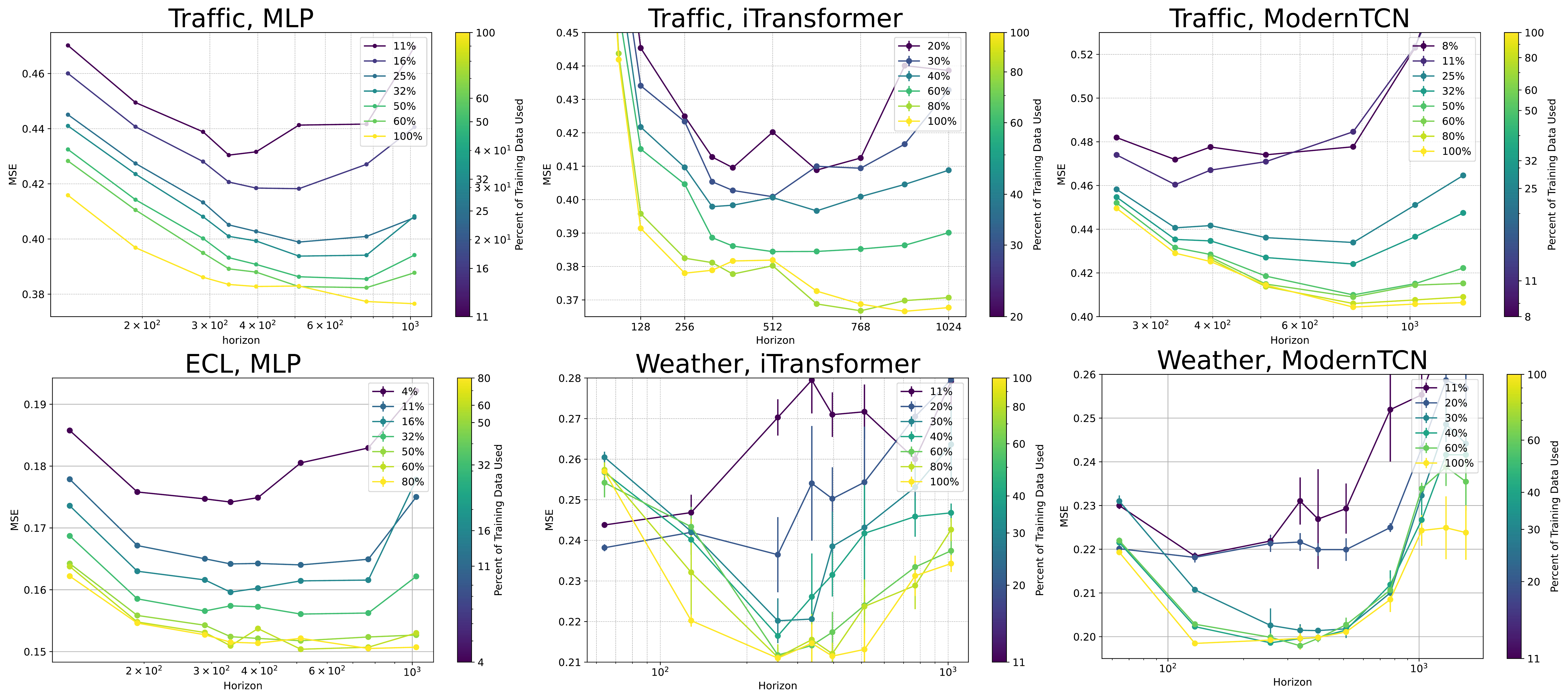}
    \caption{Loss v.s. Horizon for a certain amount of training data, for different datasets and different models.}
    \label{fig:HorizonXData_diffDatasets}
\end{figure*}

\textbf{Optimal Horizon grows with more available training data}.
We conduct experiments, fixing the available training data amount and model size while adjusting the horizon. As shown in Figure \ref{fig:HorizonXData_diffDatasets}, an optimal horizon is observed at each data amount, and this optimal value grows with more available training data.

\begin{figure*}[!h]
    \centering
    \includegraphics[width=\linewidth]{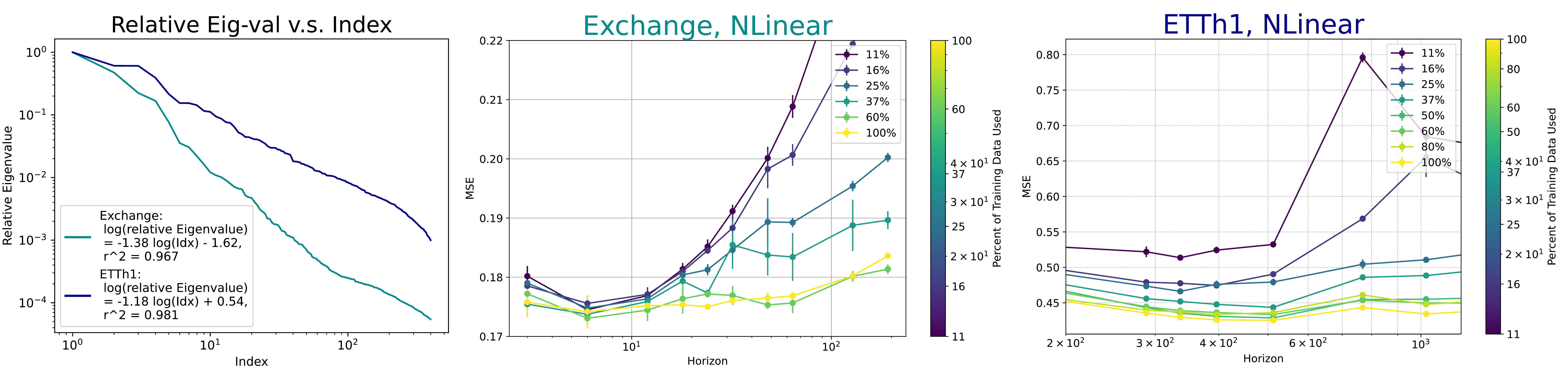}
    \caption{PCA results under Channel-Independent and Instance Normalization setting(left), Loss v.s. Horizon for certaim amount of training data on Exchange(middle) and ETTh1(right). Exchange dataset has $70\%$ data points compared to ETTh1 for training. However, since its feature degradation is stronger, the optimal horizon  ($<30$) using $100\%$ of Exchange dataset is much smaller than the optimal horizon of the ETTh1 dataset ($>300$) with only $11\%$ of available training data.}
    \label{fig:ExchangeVSETTh1}
\end{figure*}

\textbf{
A dataset with a larger feature degradation has a smaller optimal horizon.}
As our theory predicts, larger $\alpha_z$ leads to a smaller optimal horizon. In the experiment in Figure \ref{fig:ExchangeVSETTh1}, we use the same Linear model on two different datasets with comparable amount of training data: the Exchange Dataset has $70\%$ available training data compared to the ETTh1 Dataset. We use eigenvalues obtained by doing Principal component analysis on sampled series as an approximation to the feature variance in the intrinsic space. Detailed procedure and more PCA results can be found at Appendix \ref{app:pca}.


\textbf{Channel-Dependent and Channel-Independent Models} sometimes are in different states of the two cases. For CD models, $d_I(H)$ is larger and less training data is available, hence it tends to be in the few-data limit. For CI models, $d_I(H)$ is smaller and $D$ is larger, hence it may reach the data-dense limit (where the scaling exponent for $D$ is $-1$).

In the following Figure~\ref{fig:ScalingiTransformer}, iTransformer on Traffic dataset is in the data-dense limit. For MLP on Transformer, when the horizon is small it is in the few-data limit. For the Linear model, since it is simply linear (rather than piece-wise linear), we expect it to be within the data-dense limit even for long horizons and when the dataset is relatively small (like ETTh1).

\begin{figure*}[h]
    \centering
    \includegraphics[width=1\linewidth]{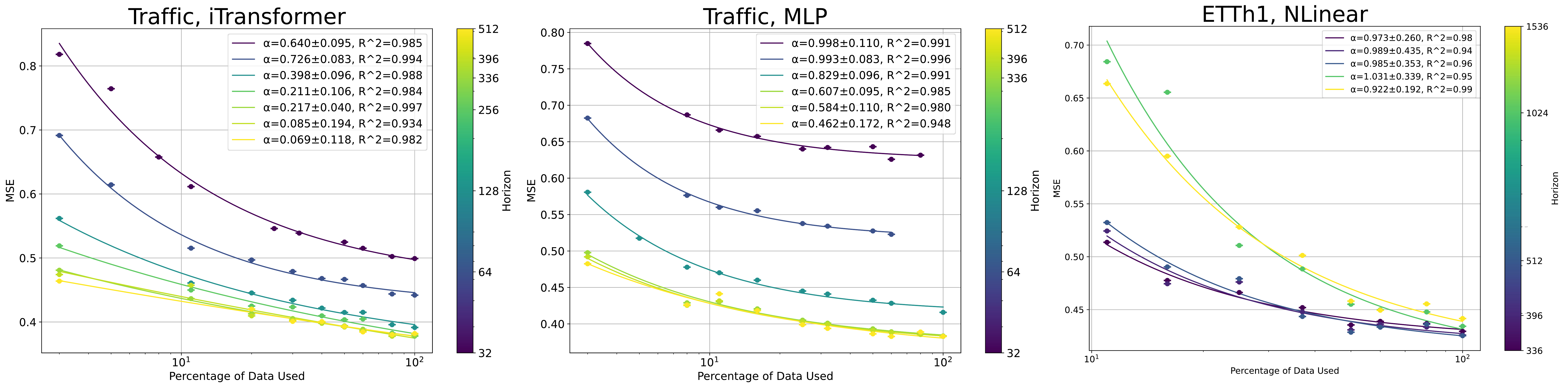}
    \caption{Data scaling behavior for iTransformer (Channel-Dependent model, left) and Norm-MLP(Channel-Independent model, middle) and NLinear (CI, right).}
    \label{fig:ScalingiTransformer}
\end{figure*}

The advantage of channel-independent and channel-dependent models can be explained from the perspective of our theory. For channel-dependent models, the horizon limitation is smaller hence it performs better with plenty of training data. For channel-independent models, more training data is available; moreover, $d$ is smaller, making the scaling exponent larger.


\subsubsection{Optimal Horizon v.s. Model Size}

As predicted by our theory there are \textbf{two cases}. \textbf{(1) When $N$ is small}, the optimal horizon does not change much with $N$. \textbf{(2) When $N$ is large}, the model size scaling term no longer dominates, the coefficient of the noise term $\frac{Nd_I}{D}$ dominants thus larger $N$ leads to smaller optimal $H$.

\begin{figure*}[!h]
    \centering
    \includegraphics[width=1\linewidth]{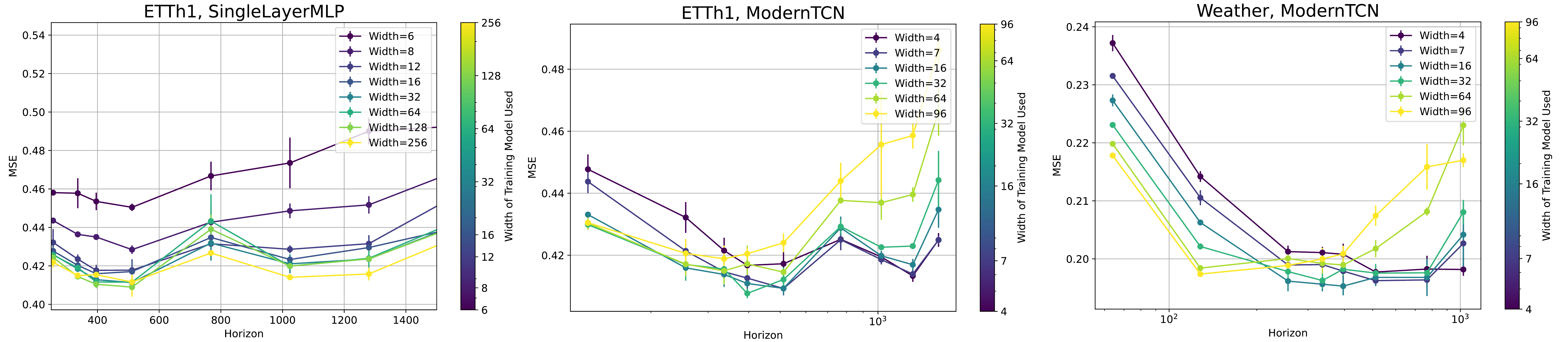}
    \caption{Loss v.s. Horizon for models of different widths. For MLP on ETTh1 (left), ModernTCN on ETTh1(middle) and ModernTCN on Weather (right).}
    \label{fig:HorizonXWidth_6}
\end{figure*}

From the observations in Figure \ref{fig:HorizonXWidth_6}, it can be discerned that in the initial scenario, where an enhancement in N results in a performance augmentation, the optimal horizon remains relatively invariant with respect to $N$. This is exemplified in the case of the SingleLayerMLP model. However, in the second scenario, where N attains its optimal value (i.e., for a certain horizon, a smaller N surpasses a larger $N$ in performance), a larger $N$ will correspond to a reduced optimal horizon. This phenomenon is evident in the instances involving the ModernTCN model.

As predicted by our theory, we see that dataset size has an impact on optimal horizon, while model size has a less significant influence on it.

\section{Discussion and Conclusion}\label{Conclusion}

\subsection{Limitation and Discussion}

Our theory mainly covers the part of time series forecasting, and our experiments verify our proposed theory on some of the well-used datasets of various sizes. However, these datasets are still small compared to some of the recently proposed large datasets\cite{liu2024timer}. The scaling behavior predicted in our work of the horizon on these large datasets remains to be experimentally verified and studied. Moreover, we mainly use popular time series forecasting models in recent work and these models might be over-designed for a few datasets for time series.

Moreover, our work mainly focus on the in-domain setting, rather for pretrained-then-finetuned models or foundation models trained on mixed datasets. We discuss more about whether the assumptions made and our theory deduction work for mixed datasets at Appendix \ref{app: foundation models}.

Since we focus on the impact of the horizon which is an adjustable hyper-parameter for forecasting tasks but a fixed hyper-parameter for other time series tasks, this work does not involve scaling behaviors for other tasks related to time series analysis.The theory for scaling law in other areas of time series forecasting as well as for foundation time series forecasting models remains to be studied. Also, although our theory is compatible with analyzing the effect of the predict length $S$ and current experiment for fixed $S$ verifies the impact of $H$, it is still worth studying the impact of $S$ on optimal $H$. Also, our theory only holds with assumptions made in Section \ref{Theory for Scaling Laws in Time Seris Forecasting}. We leave these to future work. \footnote{\small
Meanwhile, NLP and CV tasks feature long-range dependency (low degradation with horizon), large datasets, hence more detailed data granularity (e.g., context length, image resolution) produces better results in nowadays' work. However, since the community is using larger and larger models and data granularity (context length, etc), but the amount of training data is limited, it is nature to raise such question: will we see similar patterns in time series forecasting that more detailed data granularity may hurt performance in NLP or CV tasks in the future? This is also an interesting question for future work.}

\textbf{We call for future work to compare different models by `performance at optimal look back horizon' rather than `performance at a certain look back horizon'} to improve robustness. This work further elucidates that, though used in a lot of previous work, \textbf{to show a model benefits from longer horizon compared to baseline models does not necessarily prove its superiority over these baseline models}.



\subsection{Conclusion}

In this work, we propose a theoretical framework that contemplates the influence of the horizon on scaling behaviors and the performance of time series forecasting models. We take into account the size of the dataset, the complexity of the model, and the impact of the horizon on the ultimate performance. An expanded horizon results in a diminished Bayesian Error, but it simultaneously complicates the task for a limited dataset to fully encompass the entire space and for a model of restricted size to learn effectively. Furthermore, our empirical investigations corroborate the existence of scaling behaviors in relation to dataset size and model size, and validate our proposed theory concerning the impact of the horizon. 

Our theory provides some insight into the area of time series forecasting. For a certain dataset, it would be beneficial to design the models and hyperparameters according to the dataset size and feature degradation property of the particular dataset. Moreover, we think further experiments on larger foundational time series dataset about the optimal horizon with respect to pretraining loss and the loss for transferring to certain datasets may provide further insight as well.

In conclusion, we aim to provide insights to future work on time series forecasting, emphasizing the importance of the horizon and its potential impact on scaling behaviors in time series forecasting tasks.

\begin{ack}
    We would like to express our great gratitude to Professor Robert M. Haralick for significantly insightful and invaluable help and discussion. We would also like to extend our deep gratitude to Jiaye Teng, whose invaluable help and advice have been instrumental to our progress. In addition, we are profoundly grateful to Professor Yang Yuan for the enlightening discussion we had, which provided significant inspiration and insight into our work.

    We would also like to thank Professor Xiaolong Wang and Jiarui Xu at UCSD for inspiring discussions about scaling context length for NLP, and Yu Sun at Stanford for discussions about properties of time series.

    We would also like to express our gratitude to CPHOS (\url{https://cphos.cn}), an \textbf{academic non-profit organization} dedicated to providing Physics Olympiad Simulations (targeting Physics Olympiads like IPhO) for high school students for free. At CPHOS, we communicate innovative ideas and get to know great academic collaborators, without which this work could not have been carried out. 

    We would also like to thank reviewers, ACs and PCs from NeurIPS 2024 who have provided valuable suggestions which have helped to improve this work.
\end{ack}


\bibliographystyle{IEEEtran}
\bibliography{reference}

\begin{thebibliography}{10}
\providecommand{\url}[1]{#1}
\csname url@samestyle\endcsname
\providecommand{\newblock}{\relax}
\providecommand{\bibinfo}[2]{#2}
\providecommand{\BIBentrySTDinterwordspacing}{\spaceskip=0pt\relax}
\providecommand{\BIBentryALTinterwordstretchfactor}{4}
\providecommand{\BIBentryALTinterwordspacing}{\spaceskip=\fontdimen2\font plus
\BIBentryALTinterwordstretchfactor\fontdimen3\font minus \fontdimen4\font\relax}
\providecommand{\BIBforeignlanguage}[2]{{%
\expandafter\ifx\csname l@#1\endcsname\relax
\typeout{** WARNING: IEEEtran.bst: No hyphenation pattern has been}%
\typeout{** loaded for the language `#1'. Using the pattern for}%
\typeout{** the default language instead.}%
\else
\language=\csname l@#1\endcsname
\fi
#2}}
\providecommand{\BIBdecl}{\relax}
\BIBdecl

\bibitem{dlinear}
A.~Zeng, M.~Chen, L.~Zhang, and Q.~Xu, ``Are transformers effective for time series forecasting?'' 2022.

\bibitem{tsmixer}
S.-A. Chen, C.-L. Li, N.~Yoder, S.~O. Arik, and T.~Pfister, ``Tsmixer: An all-mlp architecture for time series forecasting,'' 2023.

\bibitem{xu2024fits}
Z.~Xu, A.~Zeng, and Q.~Xu, ``Fits: Modeling time series with $10k$ parameters,'' 2024.

\bibitem{informer}
H.~Zhou, S.~Zhang, J.~Peng, S.~Zhang, J.~Li, H.~Xiong, and W.~Zhang, ``Informer: Beyond efficient transformer for long sequence time-series forecasting,'' 2021.

\bibitem{autoformer}
H.~Wu, J.~Xu, J.~Wang, and M.~Long, ``Autoformer: Decomposition transformers with auto-correlation for long-term series forecasting,'' 2022.

\bibitem{fedformer}
T.~Zhou, Z.~Ma, Q.~Wen, X.~Wang, L.~Sun, and R.~Jin, ``Fedformer: Frequency enhanced decomposed transformer for long-term series forecasting,'' 2022.

\bibitem{Yuqietal-2023-PatchTST}
Y.~Nie, N.~H.~Nguyen, P.~Sinthong, and J.~Kalagnanam, ``A time series is worth 64 words: Long-term forecasting with transformers,'' in \emph{International Conference on Learning Representations}, 2023.

\bibitem{itransformer}
Y.~Liu, T.~Hu, H.~Zhang, H.~Wu, S.~Wang, L.~Ma, and M.~Long, ``itransformer: Inverted transformers are effective for time series forecasting,'' 2024.

\bibitem{timesnet}
H.~Wu, T.~Hu, Y.~Liu, H.~Zhou, J.~Wang, and M.~Long, ``Timesnet: Temporal 2d-variation modeling for general time series analysis,'' 2023.

\bibitem{moderntcn}
\BIBentryALTinterwordspacing
L.~donghao and wang xue, ``Modern{TCN}: A modern pure convolution structure for general time series analysis,'' in \emph{The Twelfth International Conference on Learning Representations}, 2024. [Online]. Available: \url{https://openreview.net/forum?id=vpJMJerXHU}
\BIBentrySTDinterwordspacing

\bibitem{micn}
H.~Wang, J.~Peng, F.~Huang, J.~Wang, J.~Chen, and Y.~Xiao, ``Micn: Multi-scale local and global context modeling for long-term series forecasting,'' 2023.

\bibitem{openaiscaling}
J.~Kaplan, S.~McCandlish, T.~Henighan, T.~B. Brown, B.~Chess, R.~Child, S.~Gray, A.~Radford, J.~Wu, and D.~Amodei, ``Scaling laws for neural language models,'' 2020.

\bibitem{chinchillascaling}
J.~Hoffmann, S.~Borgeaud, A.~Mensch, E.~Buchatskaya, T.~Cai, E.~Rutherford, D.~de~Las~Casas, L.~A. Hendricks, J.~Welbl, A.~Clark, T.~Hennigan, E.~Noland, K.~Millican, G.~van~den Driessche, B.~Damoc, A.~Guy, S.~Osindero, K.~Simonyan, E.~Elsen, J.~W. Rae, O.~Vinyals, and L.~Sifre, ``Training compute-optimal large language models,'' 2022.

\bibitem{visionscaling}
X.~Zhai, A.~Kolesnikov, N.~Houlsby, and L.~Beyer, ``Scaling vision transformers,'' 2022.

\bibitem{explainingneural}
Y.~Bahri, E.~Dyer, J.~Kaplan, J.~Lee, and U.~Sharma, ``Explaining neural scaling laws,'' 2024.

\bibitem{manifoldscaling}
U.~Sharma and J.~Kaplan, ``A neural scaling law from the dimension of the data manifold,'' 2020.

\bibitem{analysislineartimeseries}
W.~Toner and L.~Darlow, ``An analysis of linear time series forecasting models,'' 2024.

\bibitem{rmlp}
Z.~Li, S.~Qi, Y.~Li, and Z.~Xu, ``Revisiting long-term time series forecasting: An investigation on linear mapping,'' 2023.

\bibitem{revin}
\BIBentryALTinterwordspacing
T.~Kim, J.~Kim, Y.~Tae, C.~Park, J.-H. Choi, and J.~Choo, ``Reversible instance normalization for accurate time-series forecasting against distribution shift,'' in \emph{International Conference on Learning Representations}, 2021. [Online]. Available: \url{https://openreview.net/forum?id=cGDAkQo1C0p}
\BIBentrySTDinterwordspacing

\bibitem{tcn}
S.~Bai, J.~Z. Kolter, and V.~Koltun, ``An empirical evaluation of generic convolutional and recurrent networks for sequence modeling,'' 2018.

\bibitem{tcn_2}
R.~Sen, H.-F. Yu, and I.~Dhillon, ``Think globally, act locally: A deep neural network approach to high-dimensional time series forecasting,'' 2019.

\bibitem{oldtimetheories}
\BIBentryALTinterwordspacing
E.~Parzen, ``{An Approach to Time Series Analysis},'' \emph{The Annals of Mathematical Statistics}, vol.~32, no.~4, pp. 951 -- 989, 1961. [Online]. Available: \url{https://doi.org/10.1214/aoms/1177704840}
\BIBentrySTDinterwordspacing

\bibitem{25years5}
J.~G. De~Gooijer and R.~J. Hyndman, ``25 years of time series forecasting,'' \emph{International journal of forecasting}, vol.~22, no.~3, pp. 443--473, 2006.

\bibitem{timegpt}
A.~Garza, C.~Challu, and M.~Mergenthaler-Canseco, ``Timegpt-1,'' 2024.

\bibitem{liu2024timer}
Y.~Liu, H.~Zhang, C.~Li, X.~Huang, J.~Wang, and M.~Long, ``Timer: Transformers for time series analysis at scale,'' 2024.

\bibitem{das2024decoderonlyfoundationmodel}
A.~Das, W.~Kong, R.~Sen, and Y.~Zhou, ``A decoder-only foundation model for time-series forecasting,'' 2024.

\bibitem{scalinglawslargetsm}
\BIBentryALTinterwordspacing
T.~D.~P. Edwards, J.~Alvey, J.~Alsing, N.~H. Nguyen, and B.~D. Wandelt, ``Scaling-laws for large time-series models,'' 2024. [Online]. Available: \url{https://arxiv.org/abs/2405.13867}
\BIBentrySTDinterwordspacing

\bibitem{llmzeroshottimeseries}
N.~Gruver, M.~Finzi, S.~Qiu, and A.~G. Wilson, ``Large language models are zero-shot time series forecasters,'' 2023.

\bibitem{llm4ts}
C.~Chang, W.-Y. Wang, W.-C. Peng, and T.-F. Chen, ``Llm4ts: Aligning pre-trained llms as data-efficient time-series forecasters,'' 2024.

\bibitem{onefitsall}
T.~Zhou, P.~Niu, X.~Wang, L.~Sun, and R.~Jin, ``One fits all:power general time series analysis by pretrained lm,'' 2023.

\bibitem{timellm}
M.~Jin, S.~Wang, L.~Ma, Z.~Chu, J.~Y. Zhang, X.~Shi, P.-Y. Chen, Y.~Liang, Y.-F. Li, S.~Pan, and Q.~Wen, ``Time-llm: Time series forecasting by reprogramming large language models,'' 2024.

\bibitem{gnnscaling}
J.~Liu, H.~Mao, Z.~Chen, T.~Zhao, N.~Shah, and J.~Tang, ``Neural scaling laws on graphs,'' 2024.

\bibitem{quantizationerror}
P.~Zador, ``Asymptotic quantization error of continuous signals and the quantization dimension,'' \emph{IEEE Transactions on Information Theory}, vol.~28, no.~2, pp. 139--149, 1982.

\bibitem{levi2024underlying}
N.~Levi and Y.~Oz, ``The underlying scaling laws and universal statistical structure of complex datasets,'' 2024.

\bibitem{Petrini_2023}
\BIBentryALTinterwordspacing
S.~Petrini, A.~Casas-i Muñoz, J.~Cluet-i Martinell, M.~Wang, C.~Bentz, and R.~Ferrer-i Cancho, ``Direct and indirect evidence of compression of word lengths. zipf’s law of abbreviation revisited,'' \emph{Glottometrics}, vol.~54, p. 58–87, 2023. [Online]. Available: \url{http://dx.doi.org/10.53482/2023_54_407}
\BIBentrySTDinterwordspacing

\bibitem{buzug1995characterising}
T.~M. Buzug, J.~von Stamm, and G.~Pfister, ``Characterising experimental time series using local intrinsic dimension,'' \emph{Physics Letters A}, vol. 202, no. 2-3, pp. 183--190, 1995.

\bibitem{pla}
T.~N. Chan, Z.~Li, L.~H. U, and R.~Cheng, ``Plame: Piecewise-linear approximate measure for additive kernel svm,'' \emph{IEEE Transactions on Knowledge and Data Engineering}, vol.~35, no.~10, pp. 9985--9997, 2023.

\bibitem{recoverybound}
K.~D. Ba, P.~Indyk, E.~Price, and D.~P. Woodruff, ``Lower bounds for sparse recovery,'' in \emph{Proceedings of the twenty-first annual ACM-SIAM symposium on Discrete Algorithms}.\hskip 1em plus 0.5em minus 0.4em\relax SIAM, 2010, pp. 1190--1197.

\bibitem{tfb}
X.~Qiu, J.~Hu, L.~Zhou, X.~Wu, J.~Du, B.~Zhang, C.~Guo, A.~Zhou, C.~S. Jensen, Z.~Sheng, and B.~Yang, ``Tfb: Towards comprehensive and fair benchmarking of time series forecasting methods,'' 2024.

\bibitem{pytorch}
A.~Paszke, S.~Gross, F.~Massa, A.~Lerer, J.~Bradbury, G.~Chanan, T.~Killeen, Z.~Lin, N.~Gimelshein, L.~Antiga, A.~Desmaison, A.~Köpf, E.~Yang, Z.~DeVito, M.~Raison, A.~Tejani, S.~Chilamkurthy, B.~Steiner, L.~Fang, J.~Bai, and S.~Chintala, ``Pytorch: An imperative style, high-performance deep learning library,'' 2019.

\bibitem{gatedmlp}
H.~Liu, Z.~Dai, D.~R. So, and Q.~V. Le, ``Pay attention to mlps,'' 2021.

\end{thebibliography}





\newpage
\appendix

\section{More about the intrinsic space} \label{more intrinsic}

\subsection{Relationship between intrinsic dimension and horizon} \label{relation d and h}

We claimed in section \ref{intrinsic space} that $\dih$ should be asymptotically linear to $H$. Here we give a rough explanation for this. Since the reconstruction error is bounded by a constant value irrelevant to $L$, the relative error should be $O(\frac 1 L)$. That is, the error reconstructing $y\in \mathcal{O}(L)$ from $x\in \mathcal{M}(L)$ should be $O(\frac 1 L)$. On the hand, this error can be viewed as an error caused by compressing a $d_O(L)$-dimensional vector into a $d_I(L)$-dimensional space. From a sparse recovery aspect, if the covariance matrix in $\mathcal{O}(L)$ has rank $r$, then the $d_O(L)$-dimensional vector would be equivalent with a $r$-sparse vector. \cite{recoverybound} shows that, in this case, the minimal dimension that could recover it with relative error bounded by $C$ is $\tilde{\Omega}(r)$. Since in our case $r\propto L$ due to intrinsic uncertainty, hence $d_I(L)=\Omega(L)$. And since $d_I(L)\leq d_O(L)=O(L)$, we proved that $d_I(L)=O(L)$.

\subsection{A Construction Method} \label{intrinsic construction}

We informally provide a specified method to construct $\mathcal{M}(L)$ theoretically in a recursion method. If $\mathcal{M}(0),\mathcal{M}(1),\ldots,\mathcal{M}(L-1)$ has been defined, then we define $\mathcal{M}(L)$ as follows: given a time series $x_{0,1,\ldots,L-1}$ of length $L$, we represent it by concatenating the representation of $x_{0,1,\ldots,L-2}$ in $\mathcal{M}(L-1)$ and $x_{L-1}$ in a space with dimension $dim(\mathcal{M}(L-1)+1)$. Then we find a manifold that these points lie in in this space to represent $\mathcal{M}(L)$.

\subsection{Non-linear truncation} \label{non linear trunc}

We assumed that the truncating function is close to linear projection in section \ref{intrinsic space}. But in fact the two functions might have some subtle differences. This is because for a certain feature, estimating the feature with a sequence of finite length $L$ might result in error for the feature even if each element in the sequence is accurate enough. (For example, if the feature comes directly from DFT (Discrete Fourier Transform), there might be 'leakage error' if the base frequency is not an exact multiple of the feature frequency.) This is a systematic error caused by the 'precession', where the feature's characteristic frequency mismatches the measuring frequency. 

For any sequence $S$ and let $x_{t_1:t_2} \in \mathcal{M}(t_2-t_1)$ denote the vector for the sequence $S[t1:t2]$, we can define a 'truncating function' $t_p[H_1,H_2] : \mathcal{M}(H_1)\to \mathcal{M}(H_2)$ for any $H_1 \geq H_2 > 0$ with $t_p[H_1,H_2](x_{0:H_1}) = x_{0:H_2}$. Let $t_p^{-1}[H_1,H_2]$ denote the inverse image function, which maps $x\in \mathcal{M}(H_2)$ to $\{y\mid y\in \mathcal{M}(H_1), t_p[H_1,H_2](y) = x$. Notice that this truncating function is unrelated to the specific sequence. Now, assume that $t_p[H_1,H_2]$ is smooth, then from implicit function theorem we know that for any vector $x\in \mathcal M(H_1)$ if $t_p[H_1,H_2]$ is locally differentiable in a neighborhood of $x$ then $t_p^{-1}[H_1,H_2](t_p[H_1,H_2](x))$ is a submanifold of $\mathcal{M}(H_1)$ with dimension $d_I(H_1)-d_I(H_2)$, and from Sard's theorem we know that the targets of these locally undifferentiable points have measure zero. That is:

\emph{Let $S = \{x \mid x\in \mathcal{M}(H_1), t_p^{-1}[H_1,H_2](x) \notin M^{d_I(H_1)-d_I(H_2)}\}$ where $M^d$ denotes the set of all $d$-dimensional manifolds. Let $\mu$ denote a measuring function defined in $\mathcal{M}(H_1)$, then $\mu(S) = 0$.}

Hence the impact of such `critical points' to the total loss is negligible. For simplicity, we can assume $S=\emptyset$. Then $t_p$ could be written as:
$$
t_p[H_1,H_2](x) = P[H_1,H_2]g[H_1,H_2](x),
$$
where $P[H_1,H_2]$ is the projection mapping from $\mathcal{M}(H_1)$ to $\mathcal{M}(H_2)$, and $g[H_1,H_2]$ is an invertible mapping in $\mathcal{M}(H_1)$ that maps $t_p^{-1}(y)$ to a $(d_I(H_1)-d_I(H_2))$-dimensional subspace of $\mathcal{M}(H_1)$ for each $y \in \mathcal{M}(H_2)$.

We can naturally assume that the 'precession' effect on each subsequence with a certain length $l$ is only related to $l$ and intrinsic properties of the task itself, then consider a length-$2l$ sequence, it could be viewed as two independent measuring of length-$l$ subsequences, hence this error is always bounded by $O(\frac 1 {\sqrt{L}})$. 
We can formalize this condition into:
$$
\frac{\|g[H_1,H_2](x)-x\|_2}{\|g[H_1,H_2](x)\|_2} \leq \frac{\kappa}{\sqrt{H_1}},\forall x\in \mathcal{M}(H_1),
$$
where $\kappa$ is a constant, and $\|\cdot\|_2$ represents L2-norm. Also we should have $\mathbb{E}[g(x)-x]=0$. This results in an extra term in the Bayesian loss. We deduce it again formally from $\mathbb{E}[var(t_p^{-1}[H](x))]$.
$$
\begin{aligned}
    \mathbb{E}[var(t_p^{-1}[H](x))] &= \mathbb{E}[var(g^{-1}[H]P^{-1}[H](x))] \\
    &\leq \mathbb{E}[\frac{\kappa^2}{H}\|y\|_2^2] + \mathbb{E}[var(P^{-1}(x)]\\
    &= \frac{\kappa^2}{H}\sum\limits_{0\leq i < d_I}\lambda_i + \mathbb{E}[var(P^{-1}(x)]
\end{aligned}
$$
Same to what we have done in section \ref{bayesian loss}:
$$
\mathbb{E}[var(P^{-1}(x))] = \sum\limits_{d_I(H)<=i < d_I}\lambda_i
$$
Hence:
$$
\begin{aligned}
\mathbb{E}[var(t_p^{-1}[H](x))] &\leq 
\frac{\kappa^2}{H}\sum\limits_{0\leq i < d_I}\lambda_i +  \sum\limits_{d_I(H) <= i < d_I}\lambda_i \\
&\leq \frac{\lambda_0}{\alpha_Z - 1} (\frac{\kappa^2\alpha_Z}{H} + \frac{1}{d_I(H)^{2\alpha_Z-1}})
\end{aligned}
$$
which indicates that:
$$
\text{loss} \leq K^2(1-\eta)\frac{\lambda_0}{\alpha_Z - 1} (\frac{\kappa^2\alpha_Z}{H} + \frac{1}{d_I(H)^{\alpha_Z-1}})
+ \eta \sigma_M^2S^2d_I(S)
$$
Since $\kappa$ should be small, and typically $\alpha<2$ (it holds for all our experiment results), the extra term could be neglected. Hence, it is reasonable that we assume $t_p[H_1,H_2]\approx P[H_1,H_2]$.

\subsection{Reduction of Loss into Intrinsic Space}\label{app: reduction of loss into intrinsic space}

We prove that under assumptions made in Section \ref{intrinsic space}, the loss in the orginal space can be linearly bounded by the loss in the intrinsic space.

Consider we are predicting $x[0:S]$ from $x[-H:0]$, let $y_i\in \mathcal{M}(H)$ be the intrinsic vector of $x[-H:0]$ and $y_o\in \mathcal{M}(S)$ be the intrinsic vector of $x[0:S]$ (the true intrinsic vector). If we have a model $m$ so that:

$$ \mathbb{E}_{x[-H:S]\in \mathcal{O}(H+S)}[|m(y_i)-y_o|_2] \leq L_{ins} $$

where $L_{ins}$ represents the expected error (or loss) in the intrinsic space. Then, from assumption 2 we have:

$$ \mathbb{E}_{x[-H:S]\in \mathcal{O}(H+S)}[|\phi^{-1}(m(y_i)) - \phi^{-1}(y_o)|_2] \leq K_I L_{ins} $$

and from assumption 1 we know that $\mathbb{E}_{x[0:S]\in\mathcal{O}(S)}[|\phi^{-1}(y_o)-x[0:S]|_2] \leq e(S)$. Therefore:
$$
\begin{aligned}
    L_{ori} = & \mathbb{E}_{x[-H:S]\in \mathcal{O}(H+S)}[|\phi^{-1}(m(y_i)) - x[0:S]|_2] \\
    \leq & 2\mathbb{E}_{x[-H:S]\in \mathcal{O}(H+S)} [|\phi^{-1}(m(y_i)) - \phi^{-1}(y_o)|_2] + 2\mathbb{E}_{x[0:S]\in\mathcal{O}(S)}[|\phi^{-1}(y_o)-x[0:S]|_2] \\
    \leq & 2K_IL_{ins}+2e(S).
\end{aligned}
$$
Therefore, the loss in the original space is linearly bounded by the loss in the intrinsic space. W.l.o.g we may focus on studying the loss in the intrinsic space.

\section{Details about the loss deduction} \label{loss deduction}

\subsection{Boundary}

Recall that in section \ref{approx loss} we mentioned that there are two scenarios in which the model should work in totally different ways. The boundary between the two scenarios depends not only on the dataset size $D$ but also on the model size and prediction horizon. A larger model subdivides the intrinsic space into smaller subregions, each containing fewer training samples, with the number of samples per subregion inversely related to model size, expressed as $N^{-1}$.

In time series forecasting, a unique characteristic is that a single sequence, containing multiple length-$H$ subsequences, counts as multiple samples. However, these subsequences are not fully independent due to their proximity in the intrinsic space. To quantify independence, we define a separation function $f(H)$, assumed to be proportional to $H$, meaning that a sequence of total length $L$ represents $\frac{L}{f(H)}$ independent samples. Thus, the effective number of independent samples in the dataset scales with $\frac{D}{H}$.

Combining the above analysis, we can define a hyperparameter as $\xi = \frac{D}{NH}$. If $\xi$ is large enough, then we can assume data is sufficient, and the model tends to learn a linear approximation for each subregion; if $\xi$ is small, we can assume data is scarce and the model tends to find the nearest neighbour for each test sample.

\subsection{Sufficient Data} \label{loss sufficient data}

We formally derive the total loss in the scenario where we have a large amount of data. First recall that the Bayesian loss calculated in section \ref{bayesian loss} is:
$$
L_{Bayesian} \approx K_1^2(1-\eta)\frac{\lambda_0}{(\alpha_Z - 1)d_I(H)^{\alpha_Z-1}}
+ \eta\sigma_M^2S^2d_I(S)
$$
As we assume $S$ is fixed, $\eta\sigma_M^2S^2d_I(S)$ is constant and hence could be ignored. Now let's consider the approximation loss, it should come from two sources: the limited model size would result in uncertainty in the subregions, and the limited dataset size would bring noise to the data. Let's consider these two causes separately.

\subsubsection{Model Size Limitation}

We assume that a model is separating the intrinsic space into $N$ regions, and does a linear prediction for each region. It is easy to verify that the optimal prediction from $\mathcal{M}(H)$ to $\mathcal{M}(S)$ should be first-order $K_1$-Lipschitz and second-order $K_2$-Lipschitz, too.

For any region with volume $V$, the loss could be estimated as:
$$
L = \int_V \text{d}^{d_I(H)}x |f(x)-l(x)|^2
$$
where $l$ is the predicted linear function for the region. Expanding with Taylor's series, we have $f(x) \approx l(x) + \frac 1 2 (f'(x)-f'(\mu_V))(x-\mu_V)$. From the second-order Lipschitz condition, we can conclude that:
$$
|f(x)-l(x)|^2\leq K_2^2 (x-\mu_V)^4
$$
The total loss could be bounded as follows:
$$
\text{loss} \leq K_2^2 \int f(x)(x-Q(x))^4\text{d}x
$$
where $Q(x)$ represents the center of the region that $x$ is in and $f(x)$ represents the probability density. We can find that the integral in the formula is exactly the 4th power distortion of the quantization. For an arbitrary quantization with code density $g(x)$, \cite{quantizationerror} gives an exact value of distortion equal to:
$$
\int f(x)(x-Q^*(x))^4\text{d}x = \pi^{-2}N^{-\frac 4 {\dih}}\Gamma(\frac{\dih+4}{\dih})(\Gamma(\frac{\dih+2}{2}))^{\frac 4 {\dih}}\int \frac{f(x)}{g(x)^{\frac 4 {\dih}}}\text{d}x
$$
Assume $\dih$ is sufficiently large, then $\Gamma(\frac{\dih+4}{\dih})\approx 1$. So the expression can be transformed into:
$$
\int f(x)(x-Q^*(x))^4\text{d}x \approx \frac{\dih^2N^{-\frac 4 {\dih}}}{4\pi^2}\int \frac{f(x)}{g(x)^{\frac 4 {\dih}}}\text{d}x
$$
For uniform distribution, $g(x)=1$ (as we assume $\mathcal{M}(H)=[0,1]^{d_I(H)}$. Hence:
$$
\int \frac{f(x)}{g(x)^{\frac 4 {\dih}}}\text{d}x = 1
$$
And the total loss could be bounded as:
$$
\text{loss}\leq K_2^2\frac{\dih^2N^{-\frac 4 {\dih}}}{4\pi^2}
$$
From this bound, we can see that a larger horizon implies a larger burden for the model, which might lead to worse performance for the model.

\subsubsection{Dataset Limitation}

Following the previous chapter, as we assume the mapping $f$ represented by the model is a piecewise linear function, then regionally it can be written as the usual form of the gaussian-markov linear model, assuming there are homoscedastic uncorrelated noise $\varepsilon$ for every single sample in the dataset. Now we show that this noise will cause another term of loss even compared to the optimal performance of the model. We first analyze a single region containing $D_i$ samples. Consider $X$,$Y$ from the training set:
$$
Y=X\beta+\varepsilon.
$$
Where $X$ is of shape $D_i\times d_I(H)$ and $Y,\varepsilon$ is of shape $D\times d_I(S)$.  $\beta$ is the variable to learn of shape $d_I(H)\times d_I(S)$. The BLUE estimator $\hat \beta$ should be $\hat \beta = (X^TX)^{-1}X^TY$. 

Now consider $X',Y'$ from the test set:
$$
\begin{aligned}
Y'&=X'\beta+\varepsilon'\\
L_{recon} &= E[(Y'-X'\beta)^2]\\
&= E[(\varepsilon'-X'(X^TX)^{-1}X^T\varepsilon)^2]\\
&= E[\varepsilon'^2]+E[(X'(X^TX)^{-1}X^T\varepsilon)^2]
\end{aligned}
$$
The first term goes to $0$ if the test set is sufficiently large, and it is a constant given a fixed test set, so we can simply ignore it in our analysis. Consider the second term, we can show that:
$$
\mathbb{E}[(X'(X^TX)^{-1}X^T\varepsilon)^2] = D_i^{-1}d_I(H)\varepsilon_y^2
$$
Recall the noise consisting of two sources:
$$
Y = X\beta + F[S](X^*-P_c[H]^{-1}(X)) + \varepsilon_B
$$
As we've done in section \ref{bayesian loss}, we can derive that:
$$
\begin{aligned}
\text{var}(X^*-P_c[H]^{-1}(X)) &\approx \sum\limits_{d_I(H)\leq i < d_I}\lambda_i \\
&\approx \frac{\lambda_0}{(\alpha_Z-1)d_I^{\alpha_Z-1}(H)}
\end{aligned}
$$
As $F[S]$ is $K_1$-Lipschitz, this indicates that:
$$
\varepsilon_h \leq \frac{K_1^2\lambda_0}{(\alpha_Z-1)d_I^{\alpha_Z-1}(H)}
$$
And as mentioned in section \ref{bayesian loss} $\text{var}(\varepsilon_B) = \sigma_M^2S^2d_I(S)$, hence we can bound the total loss related to $D_i$ into:
$$
\text{loss}_i \leq D_i^{-1}d_I(H)(\sigma_M^2S^2d_I(S) + \frac{K_1^2\lambda_0}{(\alpha_Z-1)d_I^{\alpha_Z-1}(H)})
$$
It is worth noticing that, although different regions have their own distributions and hence may have different eigenvalues, the loss expression is irrelevant with the eigenvalues, so we can ignore this issue. Considering the entire dataset with all regions, the total loss should be:
$$
\begin{aligned}
\text{loss} &\approx \frac{\sum\limits_{j=1}^{N}\text{loss}_i \cdot D_i}{D} \\
&= \frac{Nd_I(H)}{D}(\sigma_M^2S^2d_I(S) + \frac{K_1^2\lambda_0}{(\alpha_Z-1)d_I^{\alpha_Z-1}(H)})
\end{aligned}
$$
This indicates that, using a larger model would be more prone to being affected by noise in the dataset, so especially when the data is noisy and the dataset size is limited, our theory suggests that it would perhaps be better to use a smaller model.

\subsubsection{Total loss}

Combining the above approximation loss and the Bayesian loss, the total loss would be:
$$
\text{loss} \approx K_2^2\frac{d^2N^{-\frac 4 d}}{4\pi^2} +  K_1^2(1-\eta)\frac{\lambda_0}{\alpha_Z - 1} \frac{1}{d_I(H)^{\alpha_Z-1}} +  \frac{Nd_I(H)}{D}\sigma_M^2S^2d_I(S)
$$

\subsection{Scarce Data} \label{loss scarce data}

As the model finds the nearest neighbor, the approximation loss in this scenario should consist of two sources: the distance from the test sample to its nearest neighbor and the noise in its nearest neighbor. 
If the distribution is smooth, then locally around any point $x$, the distribution of other points can be viewed as uniform when $D$ is large.

Consider a sphere of radius $r$ around a test sample, the probability that no point in the training set lies in the ball is approximately:
$$
P(r) \approx e^{-DV(r)} = e^{-DC_dr^{\dih}}
$$
where $C_d = \frac{\pi^{\frac {\dih} 2}}{\Gamma(\dih/2+1)}$ denotes the volume of a unit sphere.

From this, we can derive that:
$$
\mathbb{E}_x[r^2]\approx \frac{(\Gamma(\frac d 2 + 1))^{\frac 2 {\dih}} \Gamma(\frac 2 {\dih})}{\pi {\dih}}D^{-\frac 2 {\dih}} f(x)^{\frac 2 {\dih}}
$$
Considering $\dih$ is large, we can get that:
$$
\mathbb{E}[r^2] \approx \frac{\dih}{4\pi}D^{-\frac 2 {\dih}}
$$
which means:
$$
\text{loss}\approx \frac{K_1^2d}{4\pi}fD^{-\frac 2 d}
$$
Combining the Bayesian loss similar to the previous scenario, the total loss should be:
$$
\text{loss} \approx K_1^2(1-\eta)\frac{\lambda_0}{\alpha_Z - 1} \frac{1}{d_I(H)^{\alpha_Z-1}} + \frac{K_1^2}{4\pi}d_I(H)D^{-\frac 2 {d_I(H)}}
$$

\section{Optimal Horizon deduction} \label{optimal horizon deduction}

Continuing from section \ref{loss deduction}, we study the optimal horizon in two scenarios.

\subsection{Sufficient Data} \label{sufficientdataoptimal}

Let's first study the case where the dataset is large, which is the case in section \ref{loss sufficient data}. 

\subsubsection{Small Model} \label{smallmodeloptimal}

If the model size is too small compared to the dataset size, i.e. $N=o(D^{\frac {d_I(H)}{d_I(H)+4}})$ (however we still assume $N$ is sufficiently large compared to other variables), then the effect of dataset size on picking the optimal horizon could be neglected. In this case, the optimal value of $d_I(H)$ is unrelated with the dataset size $D$, and should satisfy:
$$
d_I^* = (\frac{K_1^2\pi^2(1-\eta)\lambda_0 N^{\frac 4 {d_I^*}}}{K_2^2\ln N})^{\frac 1 {\alpha_Z}}
$$
Solving the formula we get the following result:
$$
d_I^* = \mathcal{W}(\frac{4}{\alpha_Z C_0^{\frac 1 {\alpha_Z}}}\ln^{1+\frac 1 {\alpha_Z}}N) \approx \frac{4}{\alpha_Z C_0^{\frac 1 {\alpha_Z}}}\ln^{1+\frac 1 {\alpha_Z}}N
$$
where $\mathcal{W}(\cdot)$ is the Lambert W function and $C_0=\frac{K_1^2\pi^2(1-\eta)\lambda_0}{K_2^2}$. 

We can see that in this case, $d_I^*$ is not only irrelevant with $D$, but also changes very little with $N$. 

\subsubsection{Large Model} \label{largemodeloptimal}

On the other hand, if we assume the model is large enough, such that $N = \omega(D^{\frac {d_I(H)}{d_I(H)+4}})$, then the noise effect would be dominant in picking the optimal $H$. In this case, the optimal value of $d_I(H)$ should be:
$$
d_I^* = (\frac{K^2(1-\eta)\lambda_0 D}{N\sigma_M^2S^2d_I(S)})^{\frac 1 {\alpha_Z}}
$$

We can see that in this case the optimal $d_I$ changes rapidly to $D$ and $N$. Moreover, the optimal horizon increases with a larger dataset size $D$ and decreases with a larger model size $N$ (this is quite counter-intuitive but it surprisingly matches our experiment result).

\subsection{Scarce Data} \label{scarcedataoptimal}

This is the case in section \ref{loss scarce data}. We can directly solve $\prac{\text{loss}}{d_I(H)}=0$, which is equivalent to:
$$
\frac{4\pi (1-\eta)\lambda_0}{{d_I^*}^{\alpha_Z}}=D^{-\frac{2}{d_I^*}}(1+\frac{2\ln D}{d_I^*})
$$
We can estimate that $d_I^*=\beta C^{\frac 1 {\alpha_Z-1}}\frac{\ln D}{\ln\ln D}$
$$
d_I^* \approx \frac{2}{\alpha_Z+1}\frac{\ln D}{\ln\ln D}
$$
where $C=4\pi(1-\eta)\lambda_0$. Substituting $d_I^*$ into the formula, we can get:
$$
(\alpha_Z-1)\beta^2 + \beta - \frac{2}{C^{\frac 1 {\alpha_Z - 1}}} = 0
$$
Hence we have:
$$
\beta = \frac{\sqrt{1+\frac{8(\alpha_Z-1)}{C^{\frac 1 {\alpha_Z-1}}}}}{2(\alpha_Z-1)}
$$
and $d_I^*$ should be:
$$
d_I^* = \frac{\sqrt{1+\frac{8(\alpha_Z-1)}{C^{\frac 1 {\alpha_Z-1}}}}}{2(\alpha_Z-1)}C^{\frac 1 {\alpha_Z-1}}\frac{\ln D}{\ln\ln D} \propto \frac{\ln D}{\ln\ln D}
$$

Compared to the first scenario, the optimal $d_I$ almost doesn't change in this scenario.

\section{Details about model's partitioning}
\label{app: partition}

As we mentioned in section \ref{approx loss}, we assume the model partitions the intrinsic space uniformly, but there should be better partitions. As \cite{quantizationerror} states the optimal partition should satisfy $g(x)\propto f(x)^{\frac {\dih}{\dih+4}}$, and if the distribution is Gaussian then the loss (only for this term) should satisfy:
$$
L_p \approx  K_2^2\frac{\lambda_0^2N^{-\frac 4 {\dih}}}{e^2\dih^{2\alpha-2}} 
$$
Replacing the corresponding term with this, we see that in the case in section \ref{largemodeloptimal}, since the noise makes the most impact, this term does not matter. However, in the case in section \ref{smallmodeloptimal}, if we assume the dataset size is infinite, then the loss is asymptotically monotonically decreasing with $\dih$. Thus, the larger the horizon we use, the better performance the model will potentially reach. 

However, experiments show that the optimal horizon always exists in this case. This might suggest that the model structure limits itself from finding the best partitioning. Hence, we expect that when we design a good enough model structure (maybe only exists theoretically) and when we have enough data, we should be able to make full use of our horizon regardless of the model size or computational resources.

\section{Experiment settings}\label{app:exp}

\subsection{Dataset}
\begin{table}[h!]
\centering
\begin{tabular}{l|c|c|c|c|c}
\hline Dataset & Dim & Pred Len & Dataset Size & Frequency & Information \\
\hline ETTh1 & 7 & $192$ & $(8545,2881,2881)$ & Hourly & Electricity \\
\hline ETTh2 & 7 & $192$ & $(8545,2881,2881)$ & Hourly & Electricity \\
\hline ETTm1 & 7 & $192$ & $(34465,11521,11521)$ & $15 \mathrm{~min}$ & Electricity \\
\hline ETTm2 & 7 & $192$ & $(34465,11521,11521)$ & $15 \mathrm{~min}$ & Electricity \\
\hline Exchange & 8 & $192$ & $(5120,665,1422)$ & Daily & Economy \\
\hline Weather & 21 & $192$ & $(36792,5271,10540)$ & $10 \mathrm{~min}$ & Weather \\
\hline ECL & 321 & $192$ & $(18317,2633,5261)$ & Hourly & Electricity \\
\hline Traffic & 862 & $192$ & $(12185,1757,3509)$ & Hourly & Transportation \\
\hline
\end{tabular}\caption{Dataset used for experiments}
\end{table}
We conduct experiments on $8$ datasets listed here. ETTh1, ETTh2, ETTm1, ETTm2~\cite{informer} contains $7$ factors of electricity transformer from July 2016 - July 2018 at a certain frequency. Exchange~\cite{autoformer} includes panel data for daily exchange rates from $8$ countries across $27$ years. Weather~\cite{autoformer} includes $21$ meteorological factors from a Weather Station. ECL~\cite{autoformer} records electricity consumption of $321$ clients. Traffic~\cite{autoformer} records hourly road occupancy rates for $2$ years in San Francisco Bay area freeways. For all experiments, the pred-len is set to $192$.

In our experiments, we do $3$ iterations for ETTh1, ETTh2, ETTm1, ETTm2, Weather and Exchange, and draw graphs with error bar equaling to the standard error of these iterations. For Weather dataset, actually the error is minor. (see Appendix \ref{app:time domain diff}), and for larger datasets including Traffic and Electricity, the error is even smaller. Therefore, we conduct $1$ iters for Traffic and ECL since they are large and will introduce minor std error.

\subsection{Drop-last issue}
The drop-last issue is reported by several researchers~\cite{xu2024fits,tfb}. That is, in some previous work evaluating the model on test set with drop\_last=True setting may cause additional errors related to test batch size. In our experiments, we use drop-last set to False to avoid this issue.

\subsection{Time-related Domain Difference in Datasets: can be neglected to a certain extent}\label{app:time domain diff}
We use the first $p$ percent of our training dataset as our dataset as available training data. This may introduce an error caused by time-index-related domain differences in these datasets. To avoid this we utilize the Instance Normalization~\cite{revin}. Our further experiments on the Weather dataset show that this error is actually minor compared to the randomness introduced in the training procedure. We use an NLinear model to forecast $192$ frames based on a look-back window of size $336$. We use $10\%$ of training data, from different slices in the training set. For each slice, we repeat the experiment for $3$ times. The result is:

\begin{table}[h!]
\centering
\begin{tabular}{l|c|c|c|c}
\hline MSE & starting point=$20\%$ & $40\%$ & $60\%$ & $80\%$\\
\hline Exp1 &  $0.223$ & $0.220$ & $0.222$ & $0.223$  \\
\hline Exp2 &  $0.221$ & $0.221$ & $0.223$ & $0.224$ \\
\hline Exp3 & $0.221$ & $0.221$ & $0.223$ & $0.221$ \\
\hline Avg & $0.222\pm 0.001$ & $0.221\pm 0.001$ & $0.223\pm 0.001$ & $0.223\pm 0.001$ \\
\hline
\end{tabular}\caption{MSE measured for weather dataset}
\end{table}

It can be seen that the variance introduced by different slicing points is very small.

\subsection{Models and Experiment settings}\label{app:models and experiment settings}

For all models we utilize instance normalization that normalizes an input sequence based on its mean and std. All the deep learning networks are implemented in PyTorch\cite{pytorch} and conducted on a cluster with NVIDIA RTX 3080, RTX 3090, RTX 4090D and A100 40GB GPUs. Although we use multiple types of GPUs, it is possible to carry out all experiments on a single GPU with greater or equal to $24\text{GB}$ of memory. Experiments conducted in this work cost about $O(10^3)$ gpu hours for RTX 3090 in total.

\subsubsection{Linear Models}

For linear models, we use the linear layer after instance normalization, and perform de-normalization after the linear layer, equivalent to the NLinear model\cite{dlinear}, for all datasets and all experiments.Batch size are chosen from $\{1024,2048,4096,8192,16384\}$, learning rate is chosen from $\{0.003, 0.001\}$ and weight decay is chosen from $\{0.0005,0.005,0.001,0.0001\}$, and we decay the learning rate with a factor of $\{0.96,0.97,0.98\}$ for at most $100$ epochs with patience at most $30$. For larger datasets including Electricity, Traffic and weather the learning rate and weight decay barely make differences to the result. For relatively small datasets includgin ETTh1/ETTh2, ETTm1/ETTm2 and Exchange, larger weight decay may improve performance.

\subsubsection{MLP}

We use two types of MLPs. For the data scaling (including the HorizonXDataset) experiments, we use the gated MLP\cite{gatedmlp}: $\text{hidden}(x)=W_1(x)\odot \text{sigmoid}(W_2(x))$ where $\odot$ is the element-wise product to improve the capability of our model. For the width scaling (and the HorizonXWidth) experiments, we use plain MLP with leaky-relu as activation function.

Gated MLP is used on Traffic, Weather and Electricity dataset. On these datasets we use $3-layer$ mlp with model dimension $768$ and hidden dimension $2*768$. Learning rate is set to $0.001$ and weight decay is set to $\{0.00001, 0.0005\}$.

For single layer mlp that we use to experiment the effect of model size scaling, we us the relu activation function and the model width denotes the hidden dimension of the single hidden layer. For different datasets, the learning rate is set to $\{0.003\}$ and the weight decay is set to $\{0.0005\}$.

Batchsize is set to $4096$ or $8192$ for these cases.

\subsubsection{iTransformer}

We follow the original codes provided by iTransformer\cite{moderntcn}.

For dataset size scaling and (Horizon X Dataset Size) case we use $4$ iTransformer encoder blocks of dimension $512$ for Traffic, Weather and Electricity. For other smaller datasets, we use $1$ blocks of dimension $192$.

For Model size scaling case we use $2$ iTransformer encoder blocks for large datasets including Traffic, Weather and ECL, and $1$ block for smaller datasets. The width indicates the model dimension as well as the feedforward dimension.

Learning rate is set to $0.003$ for small datasets and $0.0003$ for large datasets. We decay the learning rate with a factor of $0.97$ or $0.98$ every epoch, and run the training procedure for at most $100$ epochs with patience $10$ for small datasets, and $40$ epochs with patience $10$ for larger datasets including Traffic, Weather and Electricity. Batch size is chosen from $\{12,20,32\}$, mainly to fit into the gpu memory.

\subsubsection{ModernTCN}

We follow the original codes provided by ModernTCN\cite{moderntcn}. We use the recommended settings given by the authors of ModernTCN except for the horizon, the channel dimension and the training procedure. Since ModernTCN has many hyperparameters that can be flexibly tuned, we do not include other same hyperparameter here, and please refer to the original paper and cods for ModernTCN for a more detailed description. Here we only include the different hyper-parameters.

For Dataset Size Scaling and (Horizon X Dataset Size) experiments, we only modify the look back horizon. For training procedure, we use learning  rate in $\{0.00005,0.0001\}$. We train for $100$ epochs with patience $20$ for small datasets and $40$ epochs with patience $10$ for larger datasets.

For Model Size Scaling and (Horizon X Model Size) experiments, we only modify the look back horizon as well as the chanenl dimension and the depth-wise dimensions in the code.Again the learning rate is chosen from $\{0.00005,0.0001\}$, and we train for $100$ epochs with patience $20$ for small datasets and $40$ epochs with patience $10$ for larger datasets. Batchsize is chosen from $\{16,64\}$ following the original hyperparameter settings given by the author of ModernTCN.

\section{Downsampling May Improve Performance: power of patch, low-pass-filter, and downsampling}

Previous work has proven the success of patches~\cite{Yuqietal-2023-PatchTST} and low-pass-filter~\cite{xu2024fits} for time series prediction. These can be explained with our theory by making the assumption that high frequency features are the most unimportant ones: thus by filtering up high frequency features we make the model visible to the most important dimensions of the intrinsic space.

\begin{figure*}[!h]
    \centering
    \includegraphics[width=\linewidth]{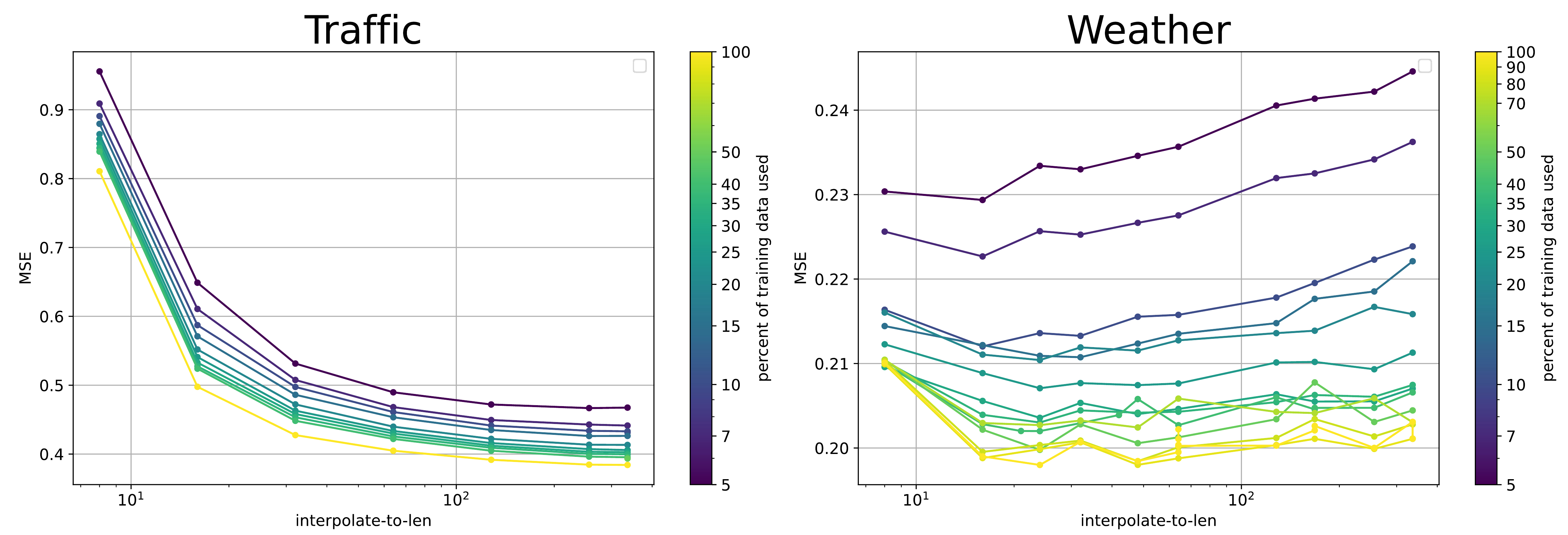}
    \caption{MSE loss v.s. interpolate-to-len (length after downsampling) for Traffic and Weather dataset for a certain amount of training data.}
    \label{fig:HorizonXWidth}
\end{figure*}

From a more detailed theoretical perspective, downsampling can be viewed as a projection to a subspace in the intrinsic space and thus has a similar effect as decreasing the horizon. Experimental results (as shown in experiments in previous works like FITS\cite{xu2024fits}, PatchTST\cite{Yuqietal-2023-PatchTST} and in Figure \ref{fig:HorizonXWidth}) show that the projected subspace of higher frequency tends to fall on the large-eigenvalue directions, or the `invisible' dimensions masked by the projection tends to be the more unimportant ones. Although the precise effect of down sampling is unknown or may need further assumptions and methods for precise consideration, it is acceptable that we may approximate the effect of down sampling to be similar to a projection to the first $d_{eff}<d_{I}(H)$ dimensions in the intrinsic space. After making this assumption, the overall loss could be expressed as:
$$
loss_{new} = L(d_{eff}) \text{where $d_{eff} < d_{I}(H)$.}
$$

Hence, if the original intrinsic dimension is larger than (local) optimal $d$, indicating $\partial loss / \partial d > 0$, reducing $d$ from $d_I(H)$ to $d_{eff}$ would help reduce loss. Otherwise if $d_I(H)$ is already smaller than (local) optimal $d^*$, meaning $\partial loss / \partial d < 0$, we would expect no performance improvement from reducing $d$ from $d_I(H)$ to $d_{eff}$.

To validate this idea, We do a similar experiment here, that is to do down-sampling and to use the downsampled sequence as input sequence. We conduct experiments on Traffic dataset and Weather dataset with a lookback horizon set to be $336$ and a prediction length set to be $192$. We use a 4-layer 512-dim gated mlp for traffic dataset and 2-layer 192-dim gated mlp for weather dataset. We can see that doing downsampling would improve the performance of the weather dataset and the optimal downsample ratio also changes with different training dataset size. The explanation for this is similar to how we explain the optimal horizon growth with the growth of the training dataset. More data in the training dataset would lead to less noise on higher dimensions, thus seeing more of these dimensions (or filtering out fewer dimensions with downsampling with longer interpolate length) can improve performance.

Meanwhile, for the traffic dataset, downsampling may hurt performance. This is because Traffic has large amount of training data, and $336$ is smaller than the optimal horizon. At this all dimensions in the intrinsic space are not dominated by noise, hence filtering them out would hurt performance, and we would expect the performance to be improved with the increment of interpolate length.
\section{PCA settings}\label{app:pca}
\begin{figure*}[!h]
    \centering
    \includegraphics[width=0.8\linewidth]{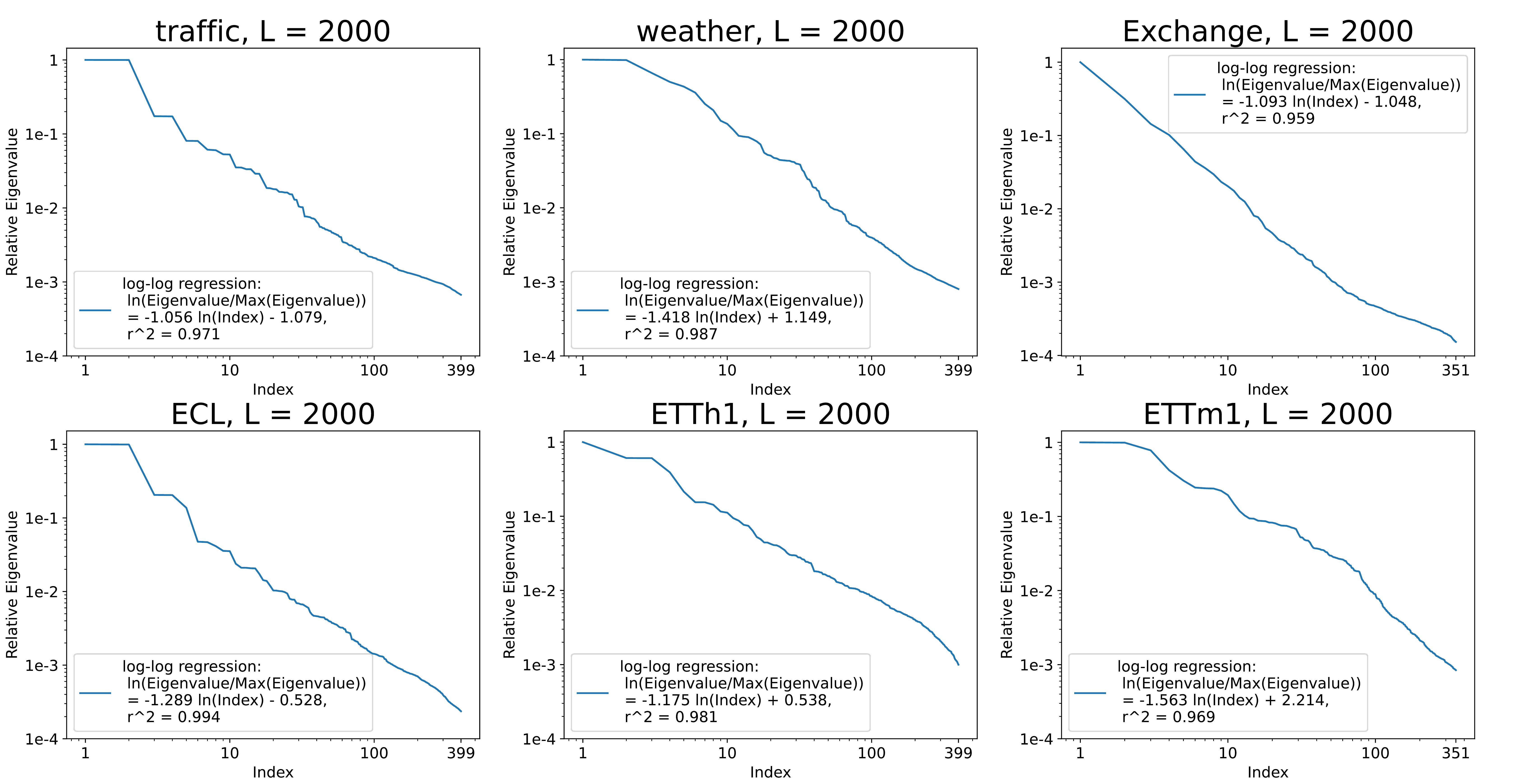}
    \caption{PCA results for $L=2000$ under Channel-Independent setting on some datasets.}
    \label{fig:PCAs_CI}
\end{figure*}

We use PCA to obtain eigen values that we use to approximate the importance of features in the intrinsic space. We conduct PCA at (all or part of the) training samples (after instance normalization), mainly for the Channel-Independent case. For the PCA used in Figure\ref{fig:ExchangeVSETTh1} and Figure \ref{fig:PCAs_CI} the horizon is set to $2000$ for both datasets and we perform PCA on all the training samples. We provide more PCA results in Figure \ref{fig:PCAs_CI}: if we approximately view the PCA eig-val as the feature importance, then they do follow a Zip-f law approximately.

\begin{figure*}[h]
  \centering
  \includegraphics[width=0.8\linewidth]{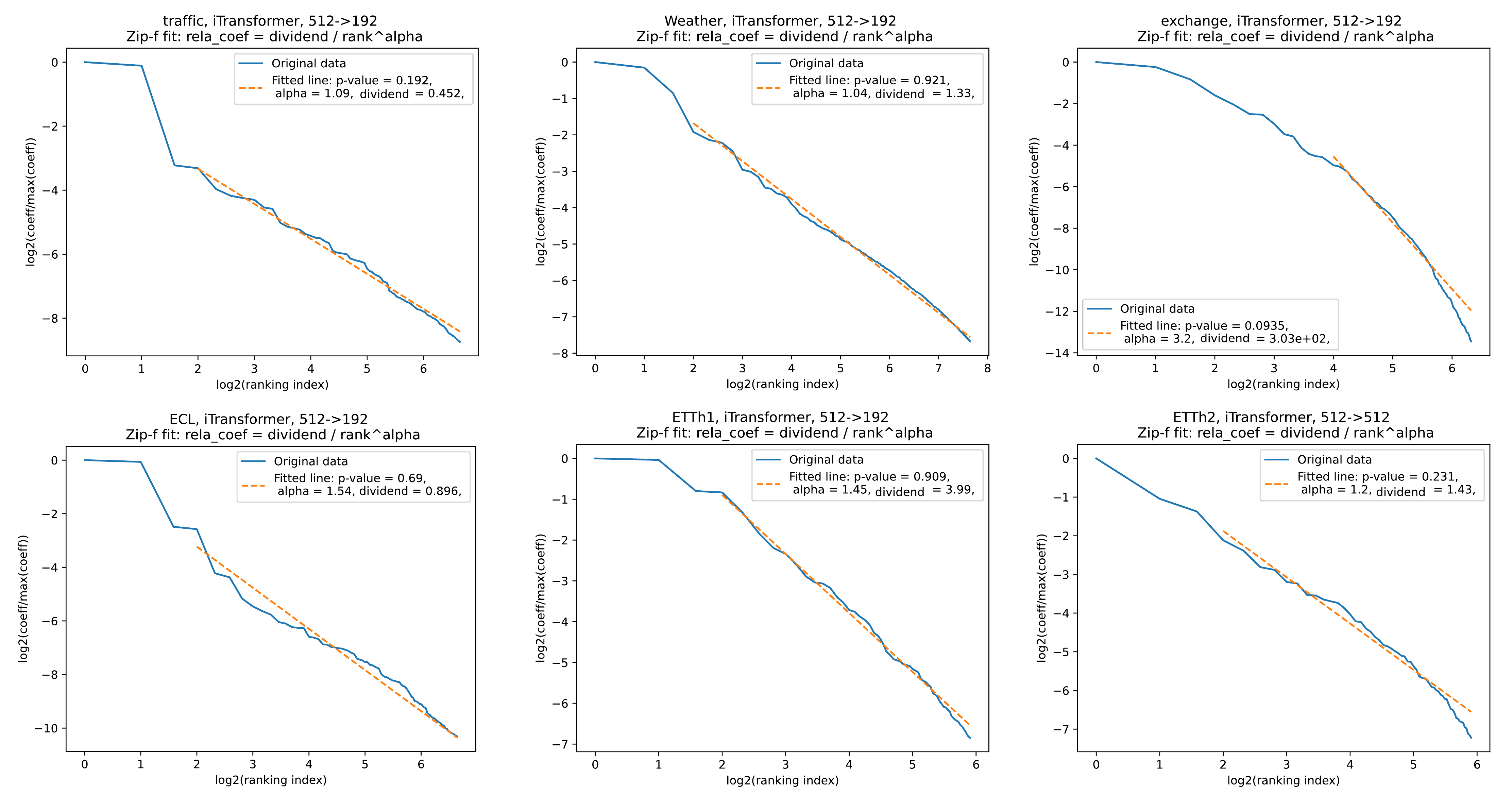}
  \caption{PCA obtained with iTransformer features. Zip-f law fitting receives $p>0.05$ in all cases. Note that it is hard for deep-learning methods (like iTransformer) to learn feature with small covariance very well, it may be hard to learn features in exchange dataset well (in which feature degrades very fast), hence the seemingly not-fitting figure for exchange is more likely caused by under-training of deep learning models.}
  \label{fig:PCA for iTransformer}
\end{figure*}

We further conduct PCA on intermediate vectors of the iTransformer\cite{itransformer} model and obtain results, further validating the Zip-f assumption in our paper for intrinsic space for Non-Linear Channel-Dependent Multivariable cases, as shown in Figure \ref{fig:PCA for iTransformer}.

\section{Foundation Models for Mixed Datasets}\label{app: foundation models}

Our theory holds without the need for any modification as long as the dataset itself follows a Zip-f distribution in the intrinsic space. This assumption is sort of natural given that Zip-f law is a natural distribution, and previous datasets we examined (like Traffic, ETT, etc) to follow the Zip-f law are composed of smaller datasets. Moreover, we provide further analysis from Theoretical and Experimental perspective on why it holds for mixed datasets.

\subsection{Zip-f distribution for Mixed Datasets: a toy model}

Theoretically, if a large dataset is composed of $s$ sub-datasets of similar size each following the Zip-f law with degradation coefficient $\alpha_1<\alpha_2<\ldots<\alpha_s$, each with size $S_i$ and follows Zip-f law: $\lambda_{ij} = A_i/j^{\alpha_i}$ where $\lambda_{ij}$ represents the $j$-th largest eigenvalue of the $i$-th dataset. Suppose the intrinsic dimensions are orthogonal with each other (hence PCA components are orthogonal). A simple assumption is that the new intrinsic space is a direct product of the old intrinsic spaces, hence eigenvalues are the union of old eigenvalues. For which, an eigenvalue of value $S$ should be the $idx_{total}$-largest eigen value, in which:
$$
idx_{total} = \sum_i idx_i = \sum_i (A_i/S)^{1/\alpha_i}.
$$
When $S$ is small (or correspondingly, when $idx_{total}$ is large) this sum is dominated by small $\alpha$s, and in limitation cases the sum is dominated by $\alpha_1$ term: $idx\approx (A_1/S)^{1/\alpha_1}+C$, which is approximately a Zip-f distribution.

\subsection{Experiment results for Zip-f distribution for Mixed Datasets}

\begin{figure*}[h]
  \centering
  \includegraphics[width=0.5\linewidth]{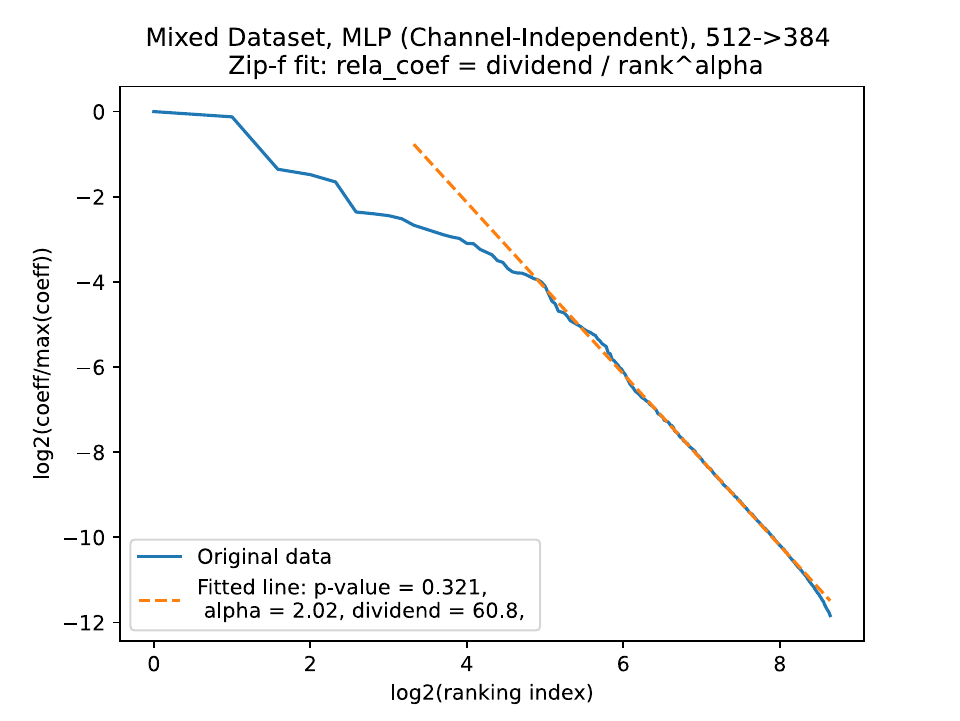}
  \caption{PCA obtained with MLP features trained on Mixed Dataset. The 2-layer 512-dimension MLP is trained under Channel Independent setting on a mixed dataset of Traffic, Weather, Exchange, ETTh1, ETTh2, ETTm1, ETTm2 and ECL. The mixed feature shows a well-aligned Zip-f law for features of higher rankings.}
  \label{fig:PCA for Mixed Datasets}
\end{figure*}

Experimentally, we use the Mixed Dataset of Traffic, Weather, ETTh1, ETTh2, ETTm1, ETTm2, exchange and ECL to train a Channel-Independent 2-layer 512-dimension MLP and use the Intermediate vector before the decoder layer as a feature vector to do PCA analysis. We found that the result follows a Zip-f distribution for higher-order components, as shown in Figure \ref{fig:PCA for Mixed Datasets}.

\section{More comparison with other formulas}\label{app:other formulas}

We compare AiC and BiC value of our proposed formula $f(x)=A+B/x^\alpha$ with other possible formulas:
\begin{itemize}
    \item $g_1(x)=A/x^\alpha$
    \item $g_2(x) = A+B\log(x)$
    \item $g_3(x) = A+Bx+Cx^2$
\end{itemize}

For ModernTCN:

\begin{table}[h!]
\centering
\begin{tabular}{lccccc}
\toprule
\textbf{AiC, BiC} & \textbf{Traffic} & \textbf{Weather} & \textbf{ETTh1} & \textbf{ETTh2} \\
\midrule
f   & \textbf{-103.9,-103.4} & \textbf{-87.2,-87.0}  & \textbf{-69.5,-69.6} & \textbf{-64.7,-65.3} \\
g1  & -95.3,-94.9            & -79.1,-79.0           & -60.6,-60.7          & -45.6,-46.0 \\
g2  & -94.6,-63.3            & -71.1,-71.0           & -59.7,-59.8          & -45.4,-45.8 \\
g3  & -93.3,-94.9            & -81.3,-83.1           & -56.5,-56.7          & -43.1,-43.8 \\
\bottomrule
\end{tabular}\caption{regression results for ModernTCN}
\end{table}

For iTransformer:

\begin{table}[h!]
\centering
\begin{tabular}{lccccc}
\toprule
\textbf{AiC, BiC} & \textbf{Traffic} & \textbf{Weather} & \textbf{ETTh1} & \textbf{ETTh2} \\
\midrule
f   & \textbf{-71.6,-70.7}  & \textbf{-74.8,-74.6}  & \textbf{-50.7,-50.9} & \textbf{-56.9,-56.7} \\
g1  & -66.7,-66.1           & -73.1,-72.9           & -45.7,-45.8          & \textbf{-57.9,-57.8} \\
g2  & -63.3,-62.6           & -71.1,-71.0           & -41.7,-41.8          & \textbf{-57.9,-57.7} \\
g3  & -54.9,-56.3           & -70.1,-72.0           & -38.9,-39.7          & \textbf{-56.4,-56.2} \\
\bottomrule
\end{tabular}\caption{regression results for iTransformer}
\end{table}

For MLP or Linear models:

\begin{table}[h!]
\centering
\begin{tabular}{lccccc}
\toprule
\textbf{AiC, BiC} & \textbf{Traffic} & \textbf{Weather} & \textbf{ETTh1} & \textbf{ETTh2} \\
\midrule
f   & \textbf{-91.9,-91.3}  & \textbf{-91.5,-91.7}  & \textbf{-89.1,-88.8} & \textbf{-62.8,-62.6} \\
g1  & -67.1,-66.7           & -83.1,-83.2           & -67.5,-67.4          & -61.1,-60.9 \\
g2  & -66.0,-65.6           & -82.1,-82.3           & -65.6,-65.5          & -59.9,-59.8 \\
g3  & -60.1,-61.2           & -81.3,-83.4           & \textbf{-89.1,-88.8} & -60.1,-59.9 \\
\bottomrule
\end{tabular}\caption{regression results for MLP or Linear models}
\end{table}

\newpage

\end{document}